%% file: scatteringStability-NeurIPS19.tex
\documentclass[preprint]{article}

\usepackage[nonatbib]{neurips_2019}

\usepackage[utf8]{inputenc} 
\usepackage[T1]{fontenc}    
\usepackage{hyperref}       
\usepackage{url}            
\usepackage{booktabs}       
\usepackage{amsfonts}       
\usepackage{nicefrac}       
\usepackage{microtype}      

\usepackage[english]{babel}
\usepackage{ifpdf}
\usepackage{cite} 
\usepackage{color}
\usepackage{pgf, tikz, pgfplots}
    \usetikzlibrary{shapes, arrows, automata, plotmarks}
    \usetikzlibrary{calc,hobby,decorations}
\usepackage[cmex10]{amsmath}
\usepackage{amssymb, amsthm}
\usepackage{mathrsfs}
\usepackage{algorithm,algpseudocode}
    \algnewcommand{\LeftComment}[1]{\Statex \(\triangleright\) #1}
\usepackage{enumerate}
\usepackage{multirow}
\usepackage{rotating}
\usepackage{subcaption}
\input{mySections.tex}

\input{mySymbol.sty}
\input{pennColors.sty}
\def\Tr{\mathsf{T}}
\def\Hr{\mathsf{H}}

\newtheorem{proposition}{\hspace{0pt}\bf Proposition}

\newtheorem{theorem}{\hspace{0pt}\bf Theorem}

\newtheorem{remark}{\hspace{0pt}\bf Remark}

\title{Stability of Graph Scattering Transforms}

\author{%
  Fernando Gama \\
  Dept. of Electrical and Systems Eng.\\
  University of Pennsylvania\\
  Philadelphia, PA 19104 \\
  \texttt{fgama@seas.upenn.edu} \\
  \And
  Joan Bruna \\
  Courant Institute of Math. Sci. \\
  New York University \\
  New York, NY 10012 \\
  \texttt{bruna@cims.nyu.edu} \\
  \And
  Alejandro Ribeiro \\
  Dept. of Electrical and Systems Eng.\\
  University of Pennsylvania\\
  Philadelphia, PA 19104 \\
  \texttt{aribeiro@seas.upenn.edu} \\
}

\begin{document}

\maketitle


\begin{abstract}
  Scattering transforms are non-trainable deep convolutional architectures that exploit the multi-scale resolution of a wavelet filter bank to obtain an appropriate representation of data. More importantly, they are proven invariant to translations, and stable to perturbations that are close to translations. This stability property dons the scattering transform with a robustness to small changes in the metric domain of the data. When considering network data, regular convolutions do not hold since the data domain presents an irregular structure given by the network topology. 
  
  In this work, we extend scattering transforms to network data by using multiresolution graph wavelets, whose computation can be obtained by means of graph convolutions. Furthermore, we prove that the resulting graph scattering transforms are stable to metric perturbations of the underlying network. This renders graph scattering transforms robust to changes on the network topology, making it particularly useful for cases of transfer learning, topology estimation or time-varying graphs.
\end{abstract}


\section{Introduction} \label{sec:intro}

\input{introductionScattering.tex}


\section{Graph scattering transforms} \label{sec:graphScattering}

\input{scatteringTransforms.tex}


\section{Stability to perturbations} \label{sec:stability}

\input{stabilityScattering.tex}


\section{Numerical results} \label{sec:sims}

\input{simsScattering.tex}


\section{Conclusions} \label{sec:conclusions}

\input{conclusionsScaterring.tex}






\bibliographystyle{IEEEtran}
\bibliography{myIEEEabrv,biblioScatteringNeurIPS}


\clearpage
\pagenumbering{arabic}
\renewcommand*{\thepage}{A\arabic{page}}
\newpage

\begin{center}
    {\Large Supplementary Materials for `Stability of Graph Scattering Transforms'}    
\end{center}

\appendix
\input{proofsScattering.tex}

\end{document}

%% file: mySections.tex
\usepackage{needspace}



%% file: introductionScattering.tex



Linear information processing architectures have been the preferred tool for extracting useful information from data due to their robustness and provable performance \cite{Rao73-LinearInference, AndersonMoore79-OptimalFiltering, Kay93-FundamentalsEstimation, Kailath80-LinearSystems, Murphy12-ProbabilisticML, Haykin96-Adaptive}. With the desire to model increasingly more complex mappings between data and useful information, linear approaches started to fall short in terms of performance, giving rise to a myriad of other nonlinear alternatives \cite[Chap. 8]{AndersonMoore79-OptimalFiltering}, \cite[Part 4]{Haykin96-Adaptive}. Of these, arguably the most successful have been convolutional neural networks (CNNs) \cite{LeCun15-DeepLearning}. CNNs consist of a cascade of layers, each of which computes a convolution with a bank of filters followed by a pointwise nonlinearity, and act as a parameterization of the nonlinear mapping between the input data and the desired useful information \cite{Goodfellow16-DeepLearning}. 

The inclusion of nonlinearities coupled with the use of trained coefficients has effectively increased the performance, but it also has obscured the limits and guarantees of CNNs \cite{Anthony99-NNtheory}. In the theoretical realm, \cite{Mallat12-Scattering, Bruna13-Scattering} opted for controlling for one of the sources of uncertainty, by fixing the bank of filters to be a set of pre-defined, multiresolution wavelets. Then, \cite{Mallat12-Scattering} proved that under admissible conditions on the wavelets, the resulting non-trainable CNN (called scattering transform) satisfies energy conservation, as well as stability to domain deformations that are close to translations. In essence, the stability properties of non-trainable scattering transforms constitutes one of the main theoretical results explaining the success of CNNs.

Data stemming from networks, however, does not exhibit a regular inherent structure that can be effectively exploited by convolutions. Data elements are, instead, related by arbitrary pairwise relationships described by an underlying graph support. Graph neural networks (GNNs) have emerged as successful architectures that exploit this graph structure \cite{Sandryhaila13-DSPG, Sandryhaila14-DSPGfreq, Shuman13-SPG, Ortega18-GSP}. GNNs, mimicking the overall architecture of CNNs, also consist of a cascade of layers, but constrain the linear transform in each layer to be a graph convolution with a bank of graph filters \cite{Scarselli09-TheGNNmodel, Bruna14-DeepSpectralNetworks, Defferrard17-CNNGraphs,Bronstein17-GeometricDeepLearning, Gama19-Architectures}. Graph convolutions are, in analogy with traditional (regular) convolutions, a weighted sum of shifted versions of the input signal. The filter taps (weights) of the bank of graph filters are also obtained by minimizing a cost function over the training set. The mathematical challenges arising from the use of trainable filters and pointwise nonlinearities have prevented a rapid development of the theory of GNNs as well. Moreover, the particularities of the underlying irregular structure supporting network data raises challenges of its own.

Following the roadmap of the Euclidean, regular case, in this paper we pursue the investigation of the benefits of GNN architectures through the lens of their non-trainable counterparts, where filters are designed from multiresolution wavelet families. Several papers \cite{ZouLerman18-Scattering, Gama19-Scattering, Perlmutter19-ScatteringManifolds} have made initial progress in defining scattering graph representation and studying their stability properties with respect to metric deformations of the domain. 
However, most of these results offer bounds that depend on the graph topology and do not hold for certain graphs or when graphs are very large. Additionally, these works do not recover the Euclidean scattering stability result on Euclidean grids. The main theoretical contribution of this work is to establish stability to \emph{relative} metric deformations for a wide class of graph wavelet families, yielding a bound that is independent on the graph topology (it only depends on the size of the deformation and the representation architecture).

The rest of the paper is structured as follows. In section~\ref{sec:related} we discuss related works. In section~\ref{sec:graphScattering} we define the scattering transform architecture, use the graph signal processing framework to describe network data (Sec.~\ref{subsec:networkData}), and define graph scattering transforms (GSTs) using graph wavelets (Sec.~\ref{subsec:graphWavelets}). Then, we proceed to prove our main theoretical claims in section~\ref{sec:stability}. Namely, that GSTs are permutation invariant (Prop.~\ref{prop:permutationInvariance}), and that they are stable (Theorem~\ref{thm:GSTstability}) under a relative perturbation model (Sec.~\ref{subsec:perturbationModel}). Finally, we show through numerical experiments in section~\ref{sec:sims}, that the GST representation is not only stable, but also captures rich enough information. Conclusions are drawn in section~\ref{sec:conclusions}.

\section{Related Work} \label{sec:related}

The particular property of stability has been investigated, in analogy to scattering transforms, for the case of non-trainable graph wavelet filter banks \cite{ZouLerman18-Scattering, Gama19-Scattering}. More specifically, \cite{ZouLerman18-Scattering} studies the stability of graph scattering transforms to permutations, as well as to perturbations on the eigenvalues and eigenvectors of the underlying graph support. Furthermore, \cite{ZouLerman18-Scattering} derives results on energy conservation. The bounds obtained on approximate permutation invariance grow with the size of the graph, while the bounds on the stability to graph perturbations are applicable only for changes in edge weights that are smaller with increasing graph size (i.e. larger graphs admit smaller edge weight changes). Alternatively, in \cite{Gama19-Scattering}, graph scattering transforms using diffusion wavelets \cite{Coifman06-DiffusionWavelets} are considered. Perturbations are defined in terms of changes in the underlying graph support, and measured using diffusion distances \cite{Nadler05-DiffusionMaps, Coifman06-DiffusionDistance}. The bounds obtained on the output for different underlying graph supports, depends on the spectral gap of the filter, making this bound quite loose in some cases \cite{Gama19-Scattering}. Finally, \cite{Levie19-Transferability} isolates the bound on the powers of the graph shift operator \cite[eq. (23)]{Gama19-Scattering} and generalizes it for arbitrary graph filters. As such, the resulting bound also depends on the spectral gap.

%% file: scatteringTransforms.tex

\begin{figure}
    \centering
    \input{figures/figGraphScatteringTransform.tex}
    \caption{Graph scattering transform. Illustration for $J=4$ scales and $L=3$ layers. At layer $\ell=0$ we have a single coefficient $\phi_{(0)}(\bbx)$ since $\ccalJ(0) = \{0\}$, which is obtained by applying the low-pass operator $U$ to the input data $\bbx$ directly. In the next layer $\ell=1$ we have $J^{1} = 4$ coefficients. We generate $4$ \emph{nodes} by applying each of the $4$ wavelets $\bbh_{j}$ to the input data followed by a pointwise nonlinearity, yielding $\bbx_{p_{j}(1)}$ where $\ccalJ(1) = \{1,2,3,4\}$. Then, we obtain the output coefficients $\phi_{p_{j}(\ell)}(\bbx)$ by means of the low-pass operator $U$. For the following layer $\ell=2$ we have $J^{2} = 16$ coefficients. For each of the $J$ previous nodes, we apply each of the wavelets yielding $J$ new nodes for each one of them, followed by the nonlinearity $\rho$. Then, we obtain the new $16$ coefficients by applying the low-pass operator $U$.}
    \label{fig:GST}
\end{figure}
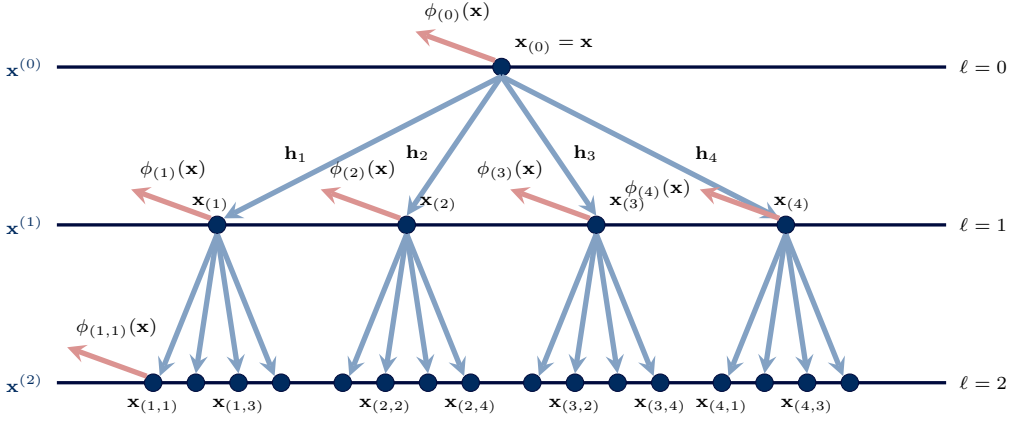

A scattering transform network \cite{Mallat12-Scattering, Bruna13-Scattering} is a deep convolutional architecture comprised of three basic elements: (i) a bank of multiresolution wavelets $\{\bbh_{j}\}_{j=1}^{J}$, (ii) a pointwise nonlinearity $\rho$ (absolute value), and (iii) a low-pass average operator $U$. These elements are combined sequentially to produce a representation $\bbPhi(\bbx)$ of the data $\bbx$. More specifically, as illustrated in Fig.~\ref{fig:GST}, each of the $J$ wavelets is applied to each of the \emph{nodes} of the previous layer, generating $J$ new \emph{nodes} to which the nonlinearity is applied. The output is harvested at each \emph{node} by computing a low-pass average through the operator $U$. For a scattering transform with $L$ layers, the number of coefficients of the representation $\bbPhi(\bbx)$ is $\sum_{\ell=0}^{L-1} J^{\ell} = (J^{L} - 1)/(J-1)$, independent of the size of the input data.

Each coefficient of the scattering transform is determined by the sequence of wavelet indices (resolution scales) traversed to compute it. We call this sequence a \emph{path}. Let $\ccalJ(\ell) = \{1,\ldots,J\}^{\ell}$ be a shorthand for the space of all possible $\ell$-tuples with $J$ elements, defined for all $\ell> 0$ and where we set $\ccalJ(0) = \{0\}$. Then, we can define the path $p_{j}(\ell): \mbN \to \ccalJ(\ell)$ as the mapping between $j \in \mbN$ and the specific sequence $p_{j}(\ell) = (j_{1},\ldots,j_{\ell})$ of length $\ell$ comprised of a combination of indices from $1$ to $J$ (tuples), with $p_{1}(0)=0$. Sequences $p_{j}(\ell)$ and $p_{i}(\ell)$ are distinct for $j \neq i$ so that $\{p_{j}(\ell)\}_{j= 1,\ldots,J^{\ell}} \equiv \ccalJ(\ell)$ is the space of all possible tuples. We denote by $\ccalJ(\ccalL) = \{p_{j}(\ell) \in \ccalJ(\ell),\forall\ j \in \{1,\ldots,J^{\ell}\}, \forall\ \ell \in \{0,\ldots,L-1\}\}$ the set of all sequences for all values of $\ell$, see Fig.~\ref{fig:GST}.

With this notation in place, the scattering transform $\bbPhi(\bbx)$ of the data $\bbx$ is the collection of \emph{scattering coefficients} $\phi_{p_{j}(\ell)}(\bbx)$
\begin{equation} \label{eqn:GST}
\bbPhi(\bbx) = 
\left\{
\phi_{p_{j}(\ell)}(\bbx)
\right\}_{\ccalJ(\ccalL)}
:= 
\left\{
\phi_{p_{j}(\ell)}(\bbx)
\right\}_{p_{j}(\ell) \in \ccalJ(\ell), \ell = 0,\ldots,L-1}.
\end{equation}
For a given sequence $p_{j}(\ell) = (j_{1},\ldots, j_{\ell}) \in \ccalJ(\ell)$, the scattering coefficient $\phi_{p_{j}(\ell)}$ is computed as
\begin{equation} \label{eqn:scatteringCoefficients}
\phi_{p_{j}(\ell)}(\bbx) =
U \left[ (\rho \bbh_{j})_{p_{j}(\ell)} \ast \bbx \right] = U \bbx_{p_{j}(\ell)}
\end{equation}
where the notation $[(\rho \bbh_{j})_{p_{j} (\ell)} \ast \bbx]: = [(\rho \bbh_{j})_{j \in p_{j}(\ell)} \ast \bbx]=\rho \bbh_{j_{\ell}} \ast \cdots \ast \rho \bbh_{j_{1}} \ast \bbx$ is a shorthand for the repeated application of pointwise nonlinearities $\rho$ and wavelets $\bbh_{j}$ following the scale indices determined by the path $p_{j}(\ell)$. The operator $U$ outputs as a scalar, computed by means of a \emph{summarizing} low-pass linear operator, typically an average or a sum. Note that we set $\phi_{p_{1}(0)} = \phi_{0} = U\bbx$. The energy of the scattering transform is given by the energy in its coefficients
\begin{equation} \label{eqn:energyScattering}
\left\| \Phi(\bbx) \right\|^{2} 
= \sum_{\ccalJ(\ccalL)} |\phi_{p_{j}(\ell)}(\bbx)|^{2}
= \sum_{\ell=0}^{L-1} \sum_{j=1}^{J^{\ell}} |\phi_{p_{j}(\ell)}(\bbx)|^{2}.
\end{equation}
%

%
%


\subsection{Network data} \label{subsec:networkData}

The scattering transform relies heavily on the use of the convolution to filter the data through the wavelet multiresolution bank. The convolution operation, in turn, depends on the data exhibiting a regular structure, such that contiguous data elements represent elements that are spatially or temporally related. This is not the case for network data, whereby data elements are related by arbitrary pairwise relationships determined by the underlying network topology.

To describe network data, we denote by $\ccalG = (\ccalV, \ccalE, \ccalW)$ the underlying graph support, with $\ccalV$ the set of $N$ nodes, $\ccalE \subseteq \ccalV \times \ccalV$ the set of edges, and $\ccalW: \ccalE \to \reals$ the edge weighing function. The data $\bbx \in \reals^{N}$ is modeled as a \emph{graph signal} where each element $[\bbx]_{i} = x_{i}$ is the value of the data at node $i \in \ccalV$\footnote{For notational simplicity, we consider that each node holds \emph{scalar} data, but the extension to \emph{vector} data is straightforward, see \cite{Defferrard17-CNNGraphs, Gama19-Architectures} for details.} \cite{Ortega18-GSP}. To operationally relate data $\bbx$ with the underlying graph support $\ccalG$, we define a \emph{graph shift operator} (GSO) $\bbS \in \reals^{N \times N}$ which is a matrix representation of the graph that respects its sparsity, i.e. $[\bbS]_{ij} = s_{ij}$ can be nonzero, only if $(j,i) \in \ccalE$ or if $i=j$ \cite{Ortega18-GSP}. Examples of GSOs commonly used in the literature include the adjacency matrix \cite{Sandryhaila13-DSPG, Sandryhaila14-DSPGfreq}, the Laplacian matrix \cite{Shuman13-SPG}, and their normalized counterparts \cite{Defferrard17-CNNGraphs, Gama19-Scattering}.

The operation $\bbS \bbx$ is, due to the sparsity constraint of $\bbS$, a local, linear operation, by which each node $i$ in the network updates its value by means of a weighted linear combination of the signal values at neighboring nodes $j \in \ccalN_{i}$
\begin{equation}
    [\bbS \bbx]_{i} = \sum_{j \in \ccalN_{i}} s_{ij} x_{j}.
\end{equation}
Note that, while $\bbS \bbx$ computes a summary of the information in the one-hop neighborhood of each node, repeated application of $\bbS$ computes summaries from further away neighborhoods, i.e. $\bbS^{k} \bbx = \bbS(\bbS^{k-1}\bbx)$ computes a summary from the $k$-hop neighborhood. This allows for the definition of \emph{graph convolutions}, in analogy with regular convolutions. More precisely, since regular convolutions are linear combinations of data that is spatially or temporally nearby, graph convolutions are defined as a linear combination of data located at consecutive neighborhoods
\begin{equation} \label{eqn:graphConvolution}
    \bbh \ast_{\bbS} \bbx = \sum_{k=0}^{K-1} h_{k} \bbS^{k} \bbx = \bbH(\bbS) \bbx
\end{equation}
where $\bbh = \{h_{0},\ldots,h_{K-1}\}$ is the set of $K$ filter coefficients, and where we use $\ast_{\bbS}$ to denote a graph convolution over GSO $\bbS$ \cite{Segarra17-Linear}. We note that the output of the graph convolution is another graph signal defined over the same graph $\ccalG$ as the input $\bbx$.

The graph convolution \eqref{eqn:graphConvolution} also satisfies the convolution theorem \cite[Sec. 2.9.6]{OppenheimSchafer10-DTSP}, which states that convolution implies multiplication in frequency domain. We define the \emph{graph frequency domain} in terms of the eigendecomposition of the GSO, which we assume to be normal $\bbS = \bbV \bbLambda \bbV^{\Hr}$, where $\bbV$ is the matrix of eigenvectors which determines the frequency basis signals, and $\bbLambda$ is the diagonal matrix of eigenvalues that determines the frequency coefficients \cite{Sandryhaila14-DSPGfreq}. The graph Fourier transform (GFT) of a graph signal is defined as the projection of the graph signal onto the space of frequency basis signals $\tbx = \bbV^{\Hr} \bbx$. So, if we compute the GFT of the output of the graph convolution, we get
\begin{equation} \label{eqn:convolutionTheorem}
    \tby 
        = \bbV^{\Hr} \bby 
        = \bbV^{\Hr} \left( \bbh \ast_{\bbS} \bbx \right) 
        = \bbV^{\Hr} \sum_{k=0}^{K-1} h_{k} \bbS^{k} \bbx
        = \sum_{k=0}^{K-1} h_{k} \bbLambda^{k} \tbx
        = \diag(\tbh) \tbx = \tbh \circ \tbx
\end{equation}
where $\circ$ denotes the elementwise (Hadamard) product, yielding an multiplication of the GFT of the filter taps with the GFT of the signal. We note that the GFT $\tbh$ of the filter coefficients $\bbh$ is given by a polynomial on the eigenvalues of the graph
\begin{equation} \label{eqn:GFTfilter}
    [\tbh]_{i} = \tdh_{i} = h(\lambda_{i}) \quad \text{with} \quad h(\lambda) = \sum_{k=0}^{K-1} h_{k} \lambda^{k}.
\end{equation}
It is very interesting to remark that the GFT of the filter is characterized by the same function $h(\lambda)$, which depends on the filter coefficients, irrespective of the graph. The specific value of the frequency coefficients of the filter (and its impact on the output), however, is obtained by instantiating $h(\lambda)$ on the eigenvalues of the given graph. But $h(\lambda)$ still characterizes the GFT of the filter taps for all graphs.


\subsection{Graph wavelets and graph scattering transforms} \label{subsec:graphWavelets}

Graph wavelets are typically defined in the graph frequency domain, by specifying a specific form on the function $h(\lambda)$ \cite{Hammond11-Wavelets, Shuman15-Wavelets}. For instance, \cite{Hammond11-Wavelets} proposes to choose a mother wavelet (wave generating kernel) $h(\lambda)$ from the regular Wavelet literature and then construct all the rest of the wavelet scales by rescaling the continuous parameter $\lambda$ before sampling it with the eigenvalues corresponding to the specific graph, see \cite[eq. (65)]{Hammond11-Wavelets} for a concrete example of a graph wavelet. This same construction method is further developed in \cite{Shuman15-Wavelets} to obtain graph wavelets that are adapted to the spectrum (i.e. that localize the wavelets around the actual eigenvalues of the given graph, instead of just sampling rescaled versions of the wavelets). Concrete examples of graph wavelets are given in \cite[Sec. IV-A]{Shuman15-Wavelets}.

Once the multiresolution wavelet filter bank is defined $\{h_{j}(\lambda)\}_{j=1}^{J}$ we proceed to compute the output by filtering each graph signal with the corresponding wavelet on the given graph. More precisely, consider $\bbS = \bbV \bbLambda \bbV^{\Hr}$ and define $\tbh_{j} = [h_{j}(\lambda_{1}),\ldots,h_{j}(\lambda_{N})]^{\Tr}$ by evaluating $h_{j}(\lambda)$ on each of the $N$ eigenvalues of $\bbS$. Then, we obtain [cf. \eqref{eqn:convolutionTheorem}]
\begin{equation}
    \bby_{j} = \bbV \tby_{j} = \bbV \diag(\tbh_{j}) \tbx = \bbV \diag(\tbh_{j}) \bbV^{\Hr} \bbx = \bbH_{j}(\bbS) \bbx
\end{equation}
where the output $\bby_{j}$ for each scale is computed as a linear operation $\bbH_{j}(\bbS)$ on the input data $\bbx$.

An important property of wavelets in general, and graph wavelets in particular, is that they conform a frame \cite{Shuman15-Wavelets}. This controls the spread of energy when computing the multiresolution output. For $0 < A \leq B < \infty$ and a multiresolution wavelet bank $\{\bbh_{j}\}_{j=1}^{J}$, it conforms a frame if
\begin{equation} \label{eqn:waveletFrame}
    A^{2} \|\bbx\|^{2} \leq \sum_{j=1}^{J} \|\bbH_{j}(\bbS) \bbx\|^{2} \leq B^{2} \|\bbx\|^{2}.
\end{equation}
For wavelets constructed following the above method, it is proven that they always conform a frame \cite[Theorem 5.6]{Hammond11-Wavelets}. In particular, the work in \cite{Shuman15-Wavelets} designs graph wavelets that are tight, which means that $A=B$ in \eqref{eqn:waveletFrame}.

We note that every analytic function $h(\lambda)$ can be computed in terms of a graph convolution \eqref{eqn:graphConvolution}. More precisely, an analytic function can be written in terms of a power series, but since graphs are finite, in virtue of the Cayley-Hamilton theorem \cite[Theorem 2.4.2]{HornJohnson85-MatrixAnalysis}, this power series can be written as a polynomial of degree at most $N-1$, i.e. by setting $K=N$ in \eqref{eqn:graphConvolution}. Moreover, \cite[Sec.~6]{Hammond11-Wavelets} provides a method for fast computation of the output of graph wavelets, by approximation with a polynomial of order $K \ll N$.

Finally, we define a \emph{graph scattering transform} (GST), as an architecture of the form \eqref{eqn:GST}-\eqref{eqn:scatteringCoefficients}, but where we replace regular convolutions by graph convolutions \eqref{eqn:graphConvolution} with a bank of analytic graph wavelets $\{\bbh_{j}\}_{j=1}^{J}$ that conform a frame \eqref{eqn:waveletFrame}.

%% file: figures/figGraphScatteringTransform.tex

\def \thisplotscale {0.75}

\def \unit {\thisplotscale cm}

\tikzstyle {node} = 
	[circle, 
	 draw,
	 minimum width = 0.25*\unit,
	 minimum height = 0.25*\unit,
	 anchor = center]
     
\def\colorNode{pennblue}
\def\colorEdge{penndarkestblue}
\def\colorGraphFilter{pennlightestblue}
\def\colorSummarizer{pennlightestred}
\def\colorBinding{pennmediumgray1}

\def\dist{1.5}

\def\xlbl{0.1*\dist}
\def\ylbl{0.1*\dist}

\def\myLineWidth{1.25}
\def\myArrowWidth{2}
\def\angleSummarizer{160}
\def\radiusSummarizer{1.5}
\def\shortsize{10}

\def\halfImageWidth{8*\unit}
\def\distBetweenLayers{0.35*\halfImageWidth}
\def\distNodesLayerOne{0.42*\halfImageWidth}
\def\distNodesLayerTwo{0.225*\distNodesLayerOne}

{\scriptsize \begin{tikzpicture}[x = 1*\unit, y = 1*\unit]
    
    
	\node at (0, 0) (o) {};
    \path (o) ++ (-\halfImageWidth,0)
                    node (leftEdgeLayer0) {};
    \path (o) ++ (\halfImageWidth,0)
                    node (rightEdgeLayer0) {};
    \path (o) ++ (-\halfImageWidth,-\distBetweenLayers)
                    node (leftEdgeLayer1) {};
    \path (o) ++ (\halfImageWidth,-\distBetweenLayers)
                    node (rightEdgeLayer1) {};
    \path (o) ++ (-\halfImageWidth,-2*\distBetweenLayers)
                    node (leftEdgeLayer2) {};
    \path (o) ++ (\halfImageWidth,-2*\distBetweenLayers)
                    node (rightEdgeLayer2) {};   
    
    \draw [draw=\colorEdge, line width = \myLineWidth] 
          (leftEdgeLayer0) -- (rightEdgeLayer0);
    \draw [draw=\colorEdge, line width = \myLineWidth] 
          (leftEdgeLayer1) -- (rightEdgeLayer1);
    \draw [draw=\colorEdge, line width = \myLineWidth] 
          (leftEdgeLayer2) -- (rightEdgeLayer2);
    
	\path (o) 
		      node (0) [fill = \colorNode, node, draw = \colorEdge] {};
    \path (0) ++ (-1.5*\distNodesLayerOne,-\distBetweenLayers)
              node (1) [fill = \colorNode, node, draw = \colorEdge] {};
    \path (0) ++ (-0.5*\distNodesLayerOne,-\distBetweenLayers)
              node (2) [fill = \colorNode, node, draw = \colorEdge] {};
    \path (0) ++ (0.5*\distNodesLayerOne,-\distBetweenLayers)
              node (3) [fill = \colorNode, node, draw = \colorEdge] {};
    \path (0) ++ (1.5*\distNodesLayerOne,-\distBetweenLayers)
              node (4) [fill = \colorNode, node, draw = \colorEdge] {};
    \path (1) ++ (-1.5*\distNodesLayerTwo,-\distBetweenLayers)
              node (11) [fill = \colorNode, node, draw = \colorEdge] {};
    \path (1) ++ (-0.5*\distNodesLayerTwo,-\distBetweenLayers)
              node (12) [fill = \colorNode, node, draw = \colorEdge] {};
    \path (1) ++ (0.5*\distNodesLayerTwo,-\distBetweenLayers)
              node (13) [fill = \colorNode, node, draw = \colorEdge] {};
    \path (1) ++ (1.5*\distNodesLayerTwo,-\distBetweenLayers)
              node (14) [fill = \colorNode, node, draw = \colorEdge] {};
    \path (2) ++ (-1.5*\distNodesLayerTwo,-\distBetweenLayers)
              node (21) [fill = \colorNode, node, draw = \colorEdge] {};
    \path (2) ++ (-0.5*\distNodesLayerTwo,-\distBetweenLayers)
              node (22) [fill = \colorNode, node, draw = \colorEdge] {};
    \path (2) ++ (0.5*\distNodesLayerTwo,-\distBetweenLayers)
              node (23) [fill = \colorNode, node, draw = \colorEdge] {};
    \path (2) ++ (1.5*\distNodesLayerTwo,-\distBetweenLayers)
              node (24) [fill = \colorNode, node, draw = \colorEdge] {};
    \path (3) ++ (-1.5*\distNodesLayerTwo,-\distBetweenLayers)
              node (31) [fill = \colorNode, node, draw = \colorEdge] {};
    \path (3) ++ (-0.5*\distNodesLayerTwo,-\distBetweenLayers)
              node (32) [fill = \colorNode, node, draw = \colorEdge] {};
    \path (3) ++ (0.5*\distNodesLayerTwo,-\distBetweenLayers)
              node (33) [fill = \colorNode, node, draw = \colorEdge] {};
    \path (3) ++ (1.5*\distNodesLayerTwo,-\distBetweenLayers)
              node (34) [fill = \colorNode, node, draw = \colorEdge] {};
    \path (4) ++ (-1.5*\distNodesLayerTwo,-\distBetweenLayers)
              node (41) [fill = \colorNode, node, draw = \colorEdge] {};
    \path (4) ++ (-0.5*\distNodesLayerTwo,-\distBetweenLayers)
              node (42) [fill = \colorNode, node, draw = \colorEdge] {};
    \path (4) ++ (0.5*\distNodesLayerTwo,-\distBetweenLayers)
              node (43) [fill = \colorNode, node, draw = \colorEdge] {};
    \path (4) ++ (1.5*\distNodesLayerTwo,-\distBetweenLayers)
              node (44) [fill = \colorNode, node, draw = \colorEdge] {};
              
    \path (0.south) 
       edge [draw = \colorGraphFilter, line width = \myArrowWidth, -stealth] 
       node [above left, pos = 0.666] {$\bbh_{1}$} (1.north east);
    \path (0.south) 
       edge [draw = \colorGraphFilter, line width = \myArrowWidth, -stealth] 
       node [above left, pos = 0.666] {$\bbh_{2}$} (2.north);
    \path (0.south) 
       edge [draw = \colorGraphFilter, line width = \myArrowWidth, -stealth] 
       node [above right, pos = 0.666] {$\bbh_{3}$} (3.north);
    \path (0.south) 
       edge [draw = \colorGraphFilter, line width = \myArrowWidth, -stealth] 
       node [above right, pos = 0.666] {$\bbh_{4}$} (4.north west);
    \path (1.south) 
       edge [draw = \colorGraphFilter, line width = \myArrowWidth, -stealth] 
       node {} (11.north east);
    \path (1.south) 
       edge [draw = \colorGraphFilter, line width = \myArrowWidth, -stealth] 
       node {} (12.north);
    \path (1.south) 
       edge [draw = \colorGraphFilter, line width = \myArrowWidth, -stealth] 
       node {} (13.north);
    \path (1.south) 
       edge [draw = \colorGraphFilter, line width = \myArrowWidth, -stealth] 
       node {} (14.north west);
    \path (2.south) 
       edge [draw = \colorGraphFilter, line width = \myArrowWidth, -stealth] 
       node {} (21.north east);
    \path (2.south) 
       edge [draw = \colorGraphFilter, line width = \myArrowWidth, -stealth] 
       node {} (22.north);
    \path (2.south) 
       edge [draw = \colorGraphFilter, line width = \myArrowWidth, -stealth] 
       node {} (23.north);
    \path (2.south) 
       edge [draw = \colorGraphFilter, line width = \myArrowWidth, -stealth] 
       node {} (24.north west);
    \path (3.south) 
       edge [draw = \colorGraphFilter, line width = \myArrowWidth, -stealth] 
       node {} (31.north east);
    \path (3.south) 
       edge [draw = \colorGraphFilter, line width = \myArrowWidth, -stealth] 
       node {} (32.north);
    \path (3.south) 
       edge [draw = \colorGraphFilter, line width = \myArrowWidth, -stealth] 
       node {} (33.north);
    \path (3.south) 
       edge [draw = \colorGraphFilter, line width = \myArrowWidth, -stealth] 
       node {} (34.north west);
    \path (4.south) 
       edge [draw = \colorGraphFilter, line width = \myArrowWidth, -stealth] 
       node {} (41.north east);
    \path (4.south) 
       edge [draw = \colorGraphFilter, line width = \myArrowWidth, -stealth] 
       node {} (42.north);
    \path (4.south) 
       edge [draw = \colorGraphFilter, line width = \myArrowWidth, -stealth] 
       node {} (43.north);
    \path (4.south) 
       edge [draw = \colorGraphFilter, line width = \myArrowWidth, -stealth] 
       node {} (44.north west);
    \draw [draw = \colorSummarizer, line width = \myArrowWidth, -stealth]
        (0.north west) -- + (\angleSummarizer:\radiusSummarizer)
        node [above right] {$\phi_{(0)}(\bbx)$};
    \draw [draw = \colorSummarizer, line width = \myArrowWidth, -stealth]
        (1.north west) -- + (\angleSummarizer:\radiusSummarizer)
        node [above right] {$\phi_{(1)}(\bbx)$};
    \draw [draw = \colorSummarizer, line width = \myArrowWidth, -stealth]
        (2.north west) -- + (\angleSummarizer:\radiusSummarizer)
        node [above right] {$\phi_{(2)}(\bbx)$};
    \draw [draw = \colorSummarizer, line width = \myArrowWidth, -stealth]
        (3.north west) -- + (\angleSummarizer:\radiusSummarizer)
        node [above] {$\phi_{(3)}(\bbx)$};
    \draw [draw = \colorSummarizer, line width = \myArrowWidth, -stealth]
        (4.north west) -- + (\angleSummarizer:\radiusSummarizer)
        node [left] {$\phi_{(4)}(\bbx)$};
    \draw [draw = \colorSummarizer, line width = \myArrowWidth, -stealth]
        (11.north west) -- + (\angleSummarizer:\radiusSummarizer)
        node [above right] {$\phi_{(1,1)}(\bbx)$};

    \node at (rightEdgeLayer0) [right] {$\ell=0$};
    \node at (rightEdgeLayer1) [right] {$\ell=1$};
    \node at (rightEdgeLayer2) [right] {$\ell=2$};
    \node at (leftEdgeLayer0) [left] {\color{\colorNode} $\bbx^{(0)}$};
    \node at (leftEdgeLayer1) [left] {\color{\colorNode} $\bbx^{(1)}$};
    \node at (leftEdgeLayer2) [left] {\color{\colorNode} $\bbx^{(2)}$};
    \node at (0.north east) [above right] {$\bbx_{(0)} = \bbx$};
    \node at (1.north west) [above] {$\bbx_{(1)}$};
    \node at (2.north east) [above right] {$\bbx_{(2)}$};
    \node at (3.north east) [above right] {$\bbx_{(3)}$};
    \node at (4.north east) [above] {$\bbx_{(4)}$};
    \node at (11.south) [below] {$\bbx_{(1,1)}$};
    \node at (12.south) [below] {};
    \node at (13.south) [below] {$\bbx_{(1,3)}$};
    \node at (14.south) [below] {};
    \node at (21.south) [below] {};
    \node at (22.south) [below] {$\bbx_{(2,2)}$};
    \node at (23.south) [below] {};
    \node at (24.south) [below] {$\bbx_{(2,4)}$};
    \node at (31.south) [below] {};
    \node at (32.south) [below] {$\bbx_{(3,2)}$};
    \node at (33.south) [below] {};
    \node at (34.south) [below] {$\bbx_{(3,4)}$};
    \node at (41.south) [below] {$\bbx_{(4,1)}$};
    \node at (42.south) [below] {};
    \node at (43.south) [below] {$\bbx_{(4,3)}$};
    \node at (44.south) [below] {};
    

	
\end{tikzpicture}} 

%% file: stabilityScattering.tex


Regular scattering transforms have been proven invariant to translations and stable to perturbations (or deformations) that are close to translations. That is, the difference on the scattering transform of the original data and that of the perturbed data, is proportional to the size of the perturbation. In the case of network data, we consider perturbations to the underlying graph support. More specifically, we consider a $N$-node graph $\ccalG$ with a GSO $\bbS$ and a \emph{perturbed} $N$-node graph $\widehat{\ccalG}$ with a GSO $\hbS$. The objective, then, is to prove that the GST is a stable operation under such perturbations, namely that
\begin{equation} \label{eqn:stabilityStatement}
    \left\| \bbPhi(\bbS, \bbx) - \bbPhi(\hbS, \bbx) \right\|
        \lesssim d(\bbS, \hbS)
\end{equation}
for some distance $d(\bbS, \hbS)$ measuring the size of the perturbation. Perturbations on the underlying graph support are particularly useful in cases when the graph is unknown and needs to be estimated \cite{Segarra17-TopologyID}, or when the graph changes with time \cite{Tolstaya19-Decentralized}. Note that, since the wavelet functions $h_{j}(\lambda)$ are fixed by design, then the analysis centers around how changes in the underlying graph support affect the eigenvalues which instantiate the GFT of the wavelets, and how does the function $h_{j}(\lambda)$ change its output when instantiated in different eigenvalues.

First, we consider perturbations that arise from permutations, that amount to node reorderings. Define the set of permutation matrices as
\begin{equation}
    \ccalP = \left\{ \bbP \in \{0,1\}^{N} : \bbP \bbone  = \bbone  \ , \ \bbP^{\Tr} \bbone = \bbone \right\}.
\end{equation}
Next, we show that the GST is invariant to permutations
%
%
\begin{proposition}[Permutation invariance] \label{prop:permutationInvariance}
    Let $\ccalG$ be a graph with a GSO $\bbS$, and let $\widehat{\ccalG}$ be a \emph{permuted} graph with GSO $\hbS = \bbP^{\Tr} \bbS \bbP$. Let $\bbx$ be the input data and $\hbx = \bbP^{\Tr} \bbx$ the correspondingly permuted data. Then, it holds that
    \begin{equation} \label{eqn:permutationInvariance}
        \bbPhi(\bbS, \bbx) = \bbPhi(\hbS, \hbx)
    \end{equation}
\end{proposition}
%
%
Prop.~\ref{prop:permutationInvariance} essentially states that the GST is independent of the chosen node ordering. Furthermore, it states that the GST exploits the topological symmetries present in the graph, i.e., that nodes with the same topological neighborhood yield the same output (if the value of the signal in the neighborhood is the same). In other words, different parts of the graph are distinct inasmuch as their neighborhood topologies are distinct.


\subsection{Perturbation model} \label{subsec:perturbationModel}

When considering arbitrary perturbations $\hbS$ of $\bbS$, and in light of Prop.~\ref{prop:permutationInvariance}, we need to define a distance $d(\bbS, \hbS)$ such that, when $\hbS$ is a permutation of $\bbS$, then $d(\bbS, \hbS ) = 0$. This would imply that, in the same way regular scattering transforms are invariant to translations and stable to perturbations that are close to translations, GSTs are invariant to permutations and stable to perturbations that are close to permutations. Define the set of permutations that make $\bbS$ and $\hbS$ the closest as
\begin{equation} \label{eqn:minimalPermutationSet}
    \ccalP_{0} = \argmin_{\bbP \in \ccalP} \left\| \bbP^{\Tr} \hbS \bbP - \bbS \right\|.
\end{equation}
Then, we consider the set of error matrices to be
\begin{equation} \label{eqn:errorMatrices}
    \ccalE (\bbS, \hbS) = \left\{ \bbP^{\Tr} \hbS \bbP - \bbS = \bbE^{\Hr} \bbS + \bbS \bbE \ , \ \bbP \in \ccalP_{0} \right\}.
\end{equation}
And, since matrices $\bbE \in \ccalE(\bbS, \hbS)$ measure the (relative) difference between $\bbS$ and $\hbS$ accounting for all possible permutations, then we can define the distance that we use to measure perturbations as
\begin{equation} \label{eqn:perturbationSize}
    d(\bbS, \hbS ) = \min_{\bbE \in \ccalE(\bbS, \hbS)} \| \bbE \|.
\end{equation}
Note that, indeed, if $\hbS = \bbP^{\Tr} \bbS \bbP$ is simply a permutation of $\bbS$, then $d(\bbS, \hbS) = 0$.

%
\begin{remark} \normalfont
    The perturbation model in \eqref{eqn:errorMatrices} and the consequent distance in \eqref{eqn:perturbationSize} is a \emph{relative} permutation model. Relative perturbations successfully take into account structural characteristics of the underlying graph such as sparsity, average degree, or mean edge weights. This is not the case when considering absolute perturbations, which is the model adopted in \cite{ZouLerman18-Scattering, Gama19-Scattering, Levie19-Transferability}.
\end{remark}


\subsection{Stability of graph wavelets} \label{subsec:stabilityWavelets}

Changes in the underlying graph support directly affect the output of filtering the signal with a wavelet. That is, by changing the eigenvalues $\lambda_{i}$ on which the wavelet $h(\lambda)$ is instantiated, the filter taps $\tdh_{i}$ are changed, and so does the output $\tdy_{i}$ in virtue of \eqref{eqn:convolutionTheorem}. Thus, the first necessary result is to quantify the change in the output of a wavelet filter. Given a wavelet function $h(\lambda)$ and corresponding instantiations $\bbH(\bbS)$ and $\bbH(\hbS)$, define the wavelet output difference as
\begin{equation} \label{eqn:waveletOutputDifference}
    \| \bbH(\bbS) - \bbH(\hbS) \| 
        = \inf \left\{ c \geq 0 :
            \min_{\bbP \in \ccalP} \left\| \bbH(\bbS) \bbx - \bbP \bbH (\bbP^{\Tr} \hbS \bbP) \bbP^{\Tr} \bbx \right\| \leq c \| \bbx \| 
        \right\}.
\end{equation}
We can then bound the wavelet output difference as shown next.
%
%
\begin{proposition}[Graph wavelet stability] \label{prop:waveletStability}
    Let $\ccalG$ be a graph with GSO $\bbS$ and $\widehat{\ccalG}$ be the \emph{perturbed} graph with GSO $\hbS$, such that $d(\bbS, \hbS) \leq \varepsilon/2$. Let $\bbE \in \ccalE(\bbS, \hbS)$, consider its eigendecomposition $\bbE = \bbU \bbM \bbU^{\Hr}$ where the eigenvalues in $\bbM = \diag(m_{1},\ldots,m_{N})$ are ordered such that $|m_{1}| \leq \cdots \leq |m_{N}|$, and assume that the structural constraint $\| \bbE / m_{N} - \bbI \| \leq \varepsilon$ holds. Let $h(\lambda)$ be a graph wavelet that satisfies the integral Lipschitz constraint $|\lambda h'(\lambda)| \leq C$. Then, it holds that
    \begin{equation} \label{eqn:waveletStability}
        \| \bbH (\bbS) - \bbH(\hbS) \| \leq \varepsilon C + \ccalO(\varepsilon^{2})
    \end{equation}
\end{proposition}
%
%
The bound in Prop.~\ref{prop:waveletStability} shows that the wavelet output difference is proportional to the size $\varepsilon$ of the perturbation. The structural constraint $\| \bbE / m_{N} - \bbI \|$ limits the changes in the structure of the graph, such as changes in sparsity or average degree and determines a cost for different perturbations. For instance, changing all the edge weights by the same amount does not affect the topology structure and thus $\|\bbE/m_{N} - \bbI \| = 0$. Also, while changing some edge weights by $\varepsilon/2$ satisfies the constraint, contracting some edges by $\varepsilon/2$ and dilating others in the same amount actually requires $\|\bbE/m_{N} - \bbI \| = \ccalO(1)$. Finally, adding and/or dropping edges altogether leads to $\|\bbE/m_{N} - \bbI \| = \ccalO(1)$ as well. In a way, $d(\bbS, \hbS) \leq \varepsilon/2$ limits the maximum edge weight change, while $\| \bbE/m_{N} - \bbI\| \leq \varepsilon$ limits how the edge weight changes affect the overall graph topology.

\begin{remark} \normalfont
In what follows, we consider the low-pass average operator $U$ to be independent of the graph shift operator structure $\bbS$. In particular, we choose $U$ to be a straightforward average of the representation obtained at all nodes, i.e. $U = N^{-1} \bbone^{\Tr}$. In the appendix, we offer a proof of stability for cases in which $U$ depends on $\bbS$ as well.
\end{remark}


\subsection{Stability of graph scattering transform} \label{subsec:stabilityGST}

The integral Lipschitz condition $|\lambda h'(\lambda)| \leq C$ requires the wavelet to be \emph{constant} in high-eigenvalue frequencies (i.e. for $\lambda \to \infty$, the derivative $h'(\lambda)$ has to go to $0$). This implies that information located in high-eigenvalue frequencies cannot be adequately discriminated (i.e. the output of the wavelet is the same for a broad band of the high-eigenvalue frequencies). Therefore, integral Lipschitz wavelets are stable, but not discriminative enough.

GSTs address this issue by incorporating pointwise nonlinearities. The effect of the pointwise nonlinearities is to cause a spillage of information throughout the frequency spectrum, in particular, into low-eigenvalue frequencies, which can then be discriminated in a stable fashion. Thus, GSTs are stable and discriminative information processing architectures.

To give a bound on the stability of the GST, we first derive a bound on the difference of a single GST coefficient, when computed on different graphs.
%
%
\begin{proposition}[GST coefficient stability] \label{prop:GSTcoeffStability}
    Let $\ccalG$ be a graph with GSO $\bbS$ and $\widehat{\ccalG}$ be the \emph{perturbed} graph with GSO $\hbS$, such that $d(\bbS, \hbS) \leq \varepsilon/2$. Let $\bbE \in \ccalE(\bbS, \hbS)$, consider its eigendecomposition $\bbE = \bbU \bbM \bbU^{\Hr}$ where the eigenvalues in $\bbM = \diag(m_{1},\ldots,m_{N})$ are ordered such that $|m_{1}| \leq \cdots \leq |m_{N}|$, and assume that the structural constraint $\| \bbE / m_{N} - \bbI \| \leq \varepsilon$ holds. Consider a GST with $L$ layers and $J$ wavelet scales $h_{j}(\lambda)$, each of which satisfies the integral Lipschitz constraint $|\lambda h_{j}'(\lambda)| \leq C$ and conform a frame with bounds $0 < A \leq B$ [cf. \eqref{eqn:waveletFrame}]. Then, for the coefficient $\phi_{p_{j}(\ell)}$ associated to path $p_{j}(\ell) = (j_{1},\ldots,j_{\ell})$ it holds that
    \begin{equation}
        | \phi_{p_{j}(\ell)} (\bbS, \bbx) - \phi_{p_{j}(\ell)} (\hbS, \bbx) | \leq  \varepsilon C \ell B^{\ell-1}  \| \bbx \|.
    \end{equation}
\end{proposition}
%
%
The bound in Prop.~\ref{prop:GSTcoeffStability} can be used to prove stability for the entire GST representation.
%
%
\begin{theorem}[GST stability] \label{thm:GSTstability}
    Under the conditions of Proposition~\ref{prop:GSTcoeffStability} it holds that
    \begin{equation} \label{eqn:GSTcoeffStability}
    \left\| \bbPhi(\bbS, \bbx) - \bbPhi (\hbS, \bbx) \right\| \leq 
         \frac{\varepsilon C}{B} \left( \sum_{\ell=0}^{L-1} \ell^{2} (B^{2}J)^{\ell} \right)^{1/2}
        \| \bbx\|.
    \end{equation}
\end{theorem}
%
%
First of all, we observe that the bound \eqref{eqn:GSTcoeffStability} is linear in the perturbation size $\varepsilon$, thus proving stability of the GST transform. Also, the proportionality constant depends on the characteristics of the GST architecture, but not on the underlying graph. It is linear also in the integral Lipschitz constant $C$, and depends exponentially on the upper bound of the filters $B$ and on the number of scales $J$, with the exponential factor given by the number of layers $L$.

\begin{remark} \normalfont
Theorem \ref{thm:GSTstability} provides a bound that is independent of graph properties. This in contrast to results in \cite{ZouLerman18-Scattering} and \cite{Levie19-Transferability} that depend on spectral signatures of the graph. An interesting consequence of this fact is that it makes it ready to take limits as we grow the number of nodes in the graph. There is, in fact, no limit to be taken as the bound holds for all graphs. Of particular importance is the limit of a line graph in which case we partially recover the seminal stability results for scattering transforms using regular convolutions in \cite{Mallat12-Scattering}. The difference between Theorem \ref{thm:GSTstability} and the results in \cite{Mallat12-Scattering} is our restriction that the perturbation matrix be close to an identity. This means we can perturb the line graph by dilating all edges or by contracting all edges. Dilations and contractions can be different for different nodes but we cannot have a mix of dilation and contraction in different parts of the line. This is allowed in \cite{Mallat12-Scattering} where perturbations are arbitrary diffeomorphisms. The reason for the relative weakness of the result is that \cite{Mallat12-Scattering} leverages extrinsic geometric information that is not available in an analysis that applies to arbitrary graphs.
\end{remark}

%% file: simsScattering.tex

\begin{figure}
    \centering
        \begin{subfigure}{0.32\textwidth}
        \centering
        \includegraphics[width=\textwidth]{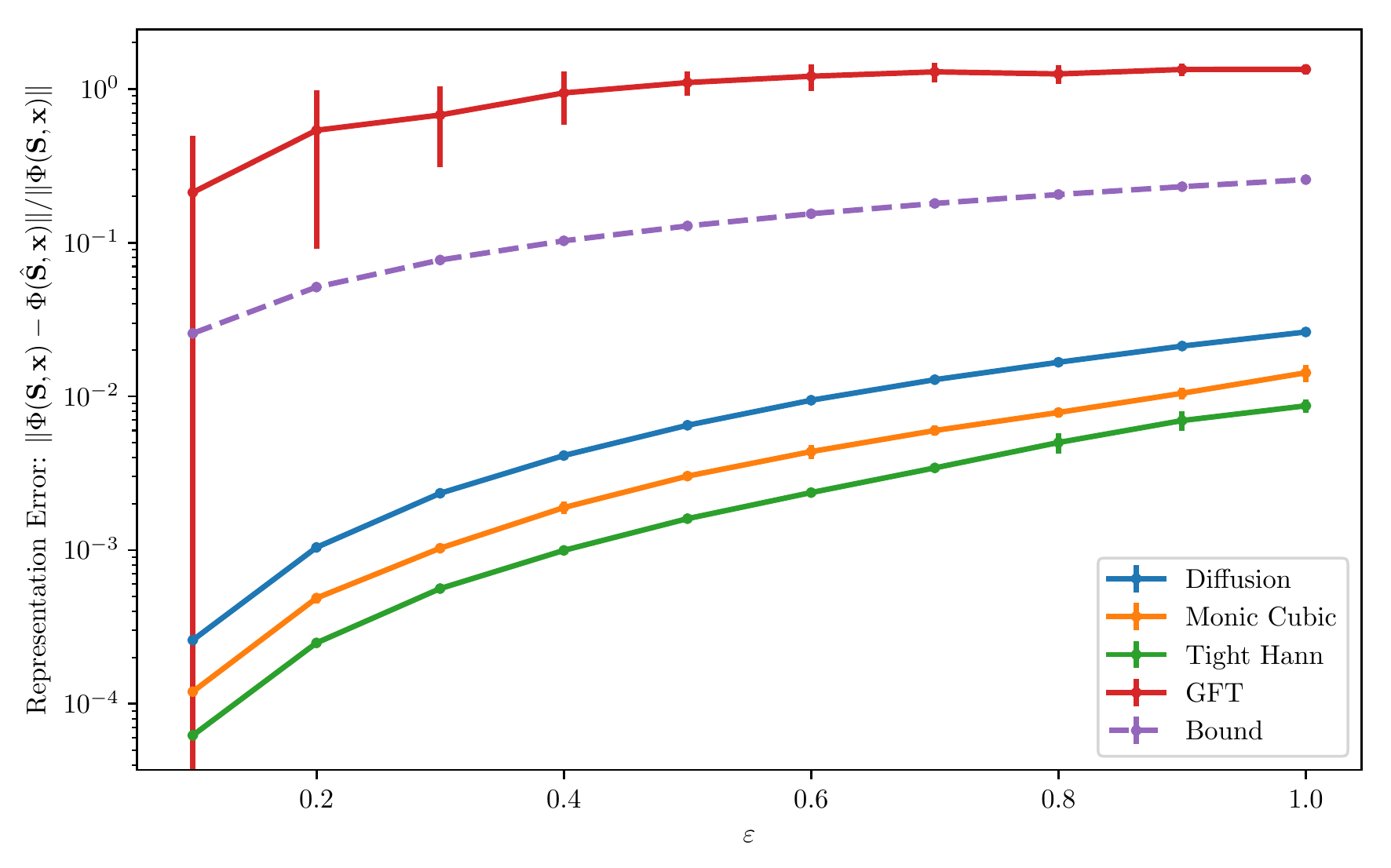}
        \caption{Small world}
        \label{sw}
    \end{subfigure}
    \hfill
    \begin{subfigure}{0.32\textwidth}
        \centering
        \includegraphics[width=\textwidth]{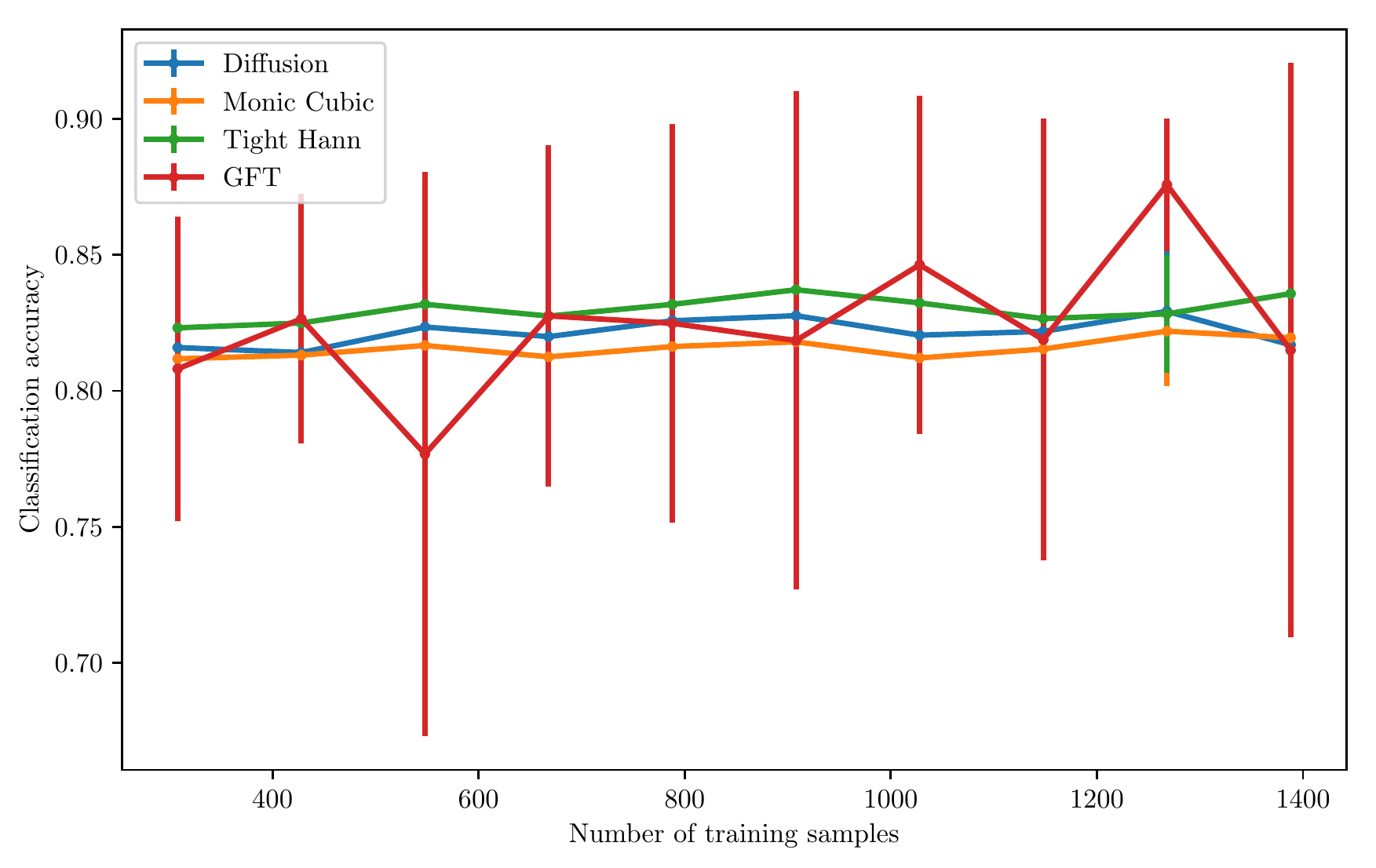}
        \caption{Authorship attribution}
        \label{wan}
    \end{subfigure}
    \hfill
    \begin{subfigure}{0.32\textwidth}
        \centering
        \includegraphics[width=\textwidth]{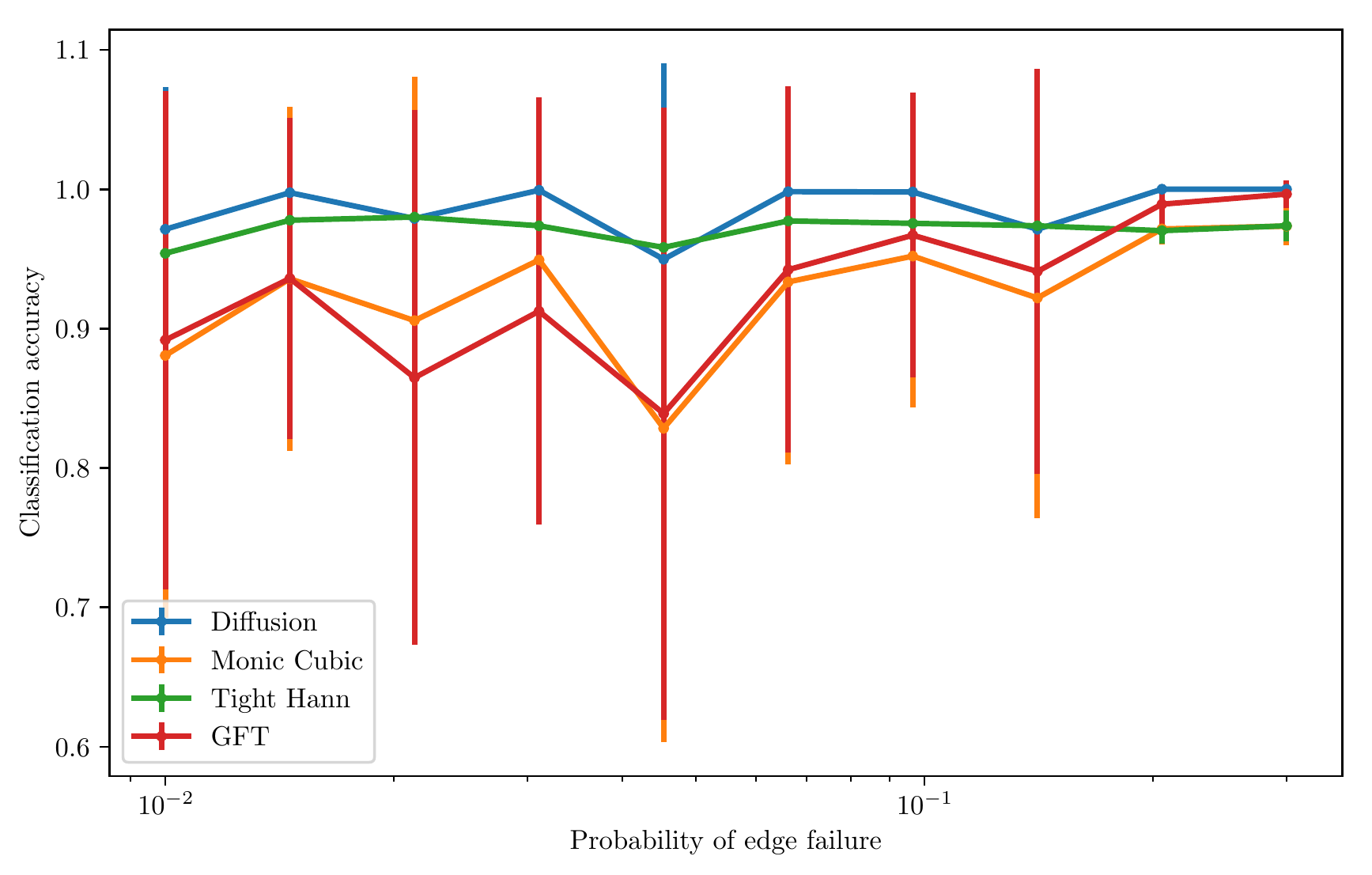}
        \caption{Facebook graph}
        \label{fb}
    \end{subfigure}
    
    \caption{\subref{sw} Difference in representation between the signal defined using the original GSO $\bbS$ and using the GSO $\hbS$ corresponding to the deformed graph as a function of the perturbation size $\varepsilon$ [cf. \eqref{eqn:perturbationSize}]. \subref{wan}-\subref{fb} Classification accuracy as a function of perturbation for the authorship attribution and the Facebook graph, respectively.}
    \label{fig:noise}
\end{figure}

For the numerical experiments, we consider three scenarios: representation error over a synthetic small world graph, authorship attribution and source localization over a Facebook subgraph, namely the same problems considered in \cite{Gama19-Scattering}. In all cases, we study the GST carried out by two different wavelets: a monic cubic polynomial as suggested in \cite{Hammond11-Wavelets} and a tight Hann wavelet as in \cite{Shuman15-Wavelets}, as well as the graph diffusion scattering introduced in \cite{Gama19-Scattering}. For comparison, we consider the GFT as a linear, graph-based representation of the data. Complete details of all simulations are provided in the appendix. We consider GSTs with $6$ scales and $3$ layers, yielding representations with $43$ coefficients; and for scenarios two and three we consider a GFTs with $43$ coefficients as well for fair comparison.

The first experiment is used to corroborate numerically the stability of the GST, and consists of computing the representation error obtained by transforming a white noise signal defined over a small world graph of $100$ nodes. We compute the relative representation error $\|\bbPhi(\bbS,\bbx)-\bbPhi(\hbS,\bbx)\|/\|\bbPhi(\bbS,\bbx)\|$ and show the results in Fig.~\ref{sw}. We observe that the GST incurs in up to $3$ orders of magnitude less relative representation error than the GFT, resulting in markedly more stable representations. Within the different choices of wavelets, the tight Hann seems to be the more stable. Also, we show the theoretical bound of Theorem \ref{thm:GSTstability} applied to the monic cubic polynomial wavelet GST, which we see is not tight, but it is still lower than the GFT.

For the second and third experiments, we consider two problems involving real-world data. The objective is twofold: (i) to show that the GST representations are, at least, as rich as the widely used GFT representation, and (ii) to consider stability to real-world perturbations (as opposed to controlled perturbations like in the first experiment). In Fig.~\ref{wan} we show the classification accuracy in a problem involving authorship attribution of texts written by Jane Austen, in the same scenario considered in \cite{Gama19-Scattering}. The perturbation comes from considering different number of training excerpts and amounts to uncertainty in estimating the underlying graph topology. It is immediate to note that the performance obtained by a linear SVM classifier operating on the GST representation is comparable to that obtained when using the GFT. We also observe that the oscillation of the mean classification accuracy of the GFT (as well as the large error bars) show that is is much less stable than the GST. In Fig.~\ref{fb} we show the classification accuracy for a source localization problem over the $234$-node Facebook subnetwork \cite{McAuley12-EgoNets}, as discussed in \cite{Gama19-Scattering}. In this case, the perturbation comes from randomly dropping edges with probability given in the x-axis of the figure (from $0.01$ to $0.3$). We observe that the GST using tight Hann wavelets and the diffusion scattering transform achieve better performance than the GFT, while the GST using monic cubic polynomials yields similar performance to the GFT. Finally, we note that the variability in the GFT is larger than in the GST, even though the variability in the GST is somehow comparable. A plausible explanation is that dropping edges in a graph incurs in a high structural cost (see Prop.~\ref{prop:GSTcoeffStability}) that degrades the stability of the GST.

%% file: conclusionsScaterring.tex

We have studied the stability properties of graph scattering transforms (GSTs) built with integral Lipschitz wavelets. We have introduced a relative perturbation model that takes into account the structure of the graph as well as its edge weights. We proved stability of the GST, by which changes in the output of the GST are bounded proportionally to size of the perturbation of the underlying graph. The proportionality constant depends on the model characteristics (number of scales, number of layers, chosen wavelets) but does not depend on characteristics of the graph. Finally, we used numerical experiments to show that the GST representation is also rich enough to achieve comparable performance as the popular GFT, which is a linear, graph-based representation.

%% file: proofsScattering.tex


\section{Proof of proposition \ref{prop:permutationInvariance}: Permutation invariance}

The low-pass summarizing operator $U$ is linear so that, under permutations, $\hhatU = U \bbP$. Therefore,
\begin{equation}
    \phi_{p_{j}(\ell)}(\hbS, \hbx) = \hhatU (\rho \bbH_{j}(\hbS))_{p_{j}(\ell)}\hbx = U \bbP (\rho \bbH_{j}(\bbP^{\Tr}\bbS \bbP))_{p_{j}(\ell)} \bbP^{\Tr}\bbx.
\end{equation}
But, for analytical wavelets, we have that $\bbH_{j}(\bbP^{\Tr} \bbS \bbP) = \bbP^{\Tr} \bbH_{j}(\bbS) \bbP$ [cf. \eqref{eqn:graphConvolution}]. Also, since the nonlinearities are pointwise, $\rho(\bbP^{\Tr} \bbz) = \bbP^{\Tr} \rho(\bbz)$ for any $\bbz$. Then, we get
\begin{equation}
    \phi_{p_{j}(\ell)}(\hbS, \hbx) = U \bbP \bbP^{\Tr} (\rho \bbH_{j}(\bbS ))_{p_{j}(\ell)} \bbP\bbP^{\Tr}\bbx = U (\rho \bbH_{j}(\bbS ))_{p_{j}(\ell)} \bbx = \phi_{p_{j}(\ell)}(\bbS, \bbx)
\end{equation}
where we used that $\bbP \bbP^{\Tr} = \bbI$ by definition of permutation matrix. Since this holds independently of $p_{j}(\ell)$, then it holds for every scattering coefficient. If it holds for every scattering coefficient, then it hold for the GST $\bbPhi(\bbx)$, thereby completing the proof.


\section{Proof of proposition \ref{prop:waveletStability}: Graph wavelet stability}

Without loss of generality assume that $\bbP = \bbI$ (alternatively, fix some $\bbP_{0} \in \ccalP$ and redefine $\hbS$ to be equal to $\bbP_{0}^{\Tr}\hbS \bbP_{0}$). Then, we can write $\hbS = \bbS + \bbE^{\Hr} \bbS + \bbS \bbE$. Observe that, given two arbitrary square matrices $\bbA$ and $\bbB$ of the same size, the first order expansion of $(\bbA + \bbB)^{k}$ gives
\begin{equation}
    (\bbA + \bbB)^{k} 
        = \bbA^{k} 
            + \sum_{r=0}^{k-1} \bbA^{r} \bbB \bbA^{k-r-1} 
            + \bbC
\end{equation}
with $\bbC$ such that $\|\bbC\| \leq \sum_{r=2}^{k} \binom{k}{r} \|\bbB\|^{r} \|\bbA\|^{k-r}$. Exploiting that the graph wavelets are analytic functions, we can use this first-order approximation in \eqref{eqn:graphConvolution} with $\bbA = \bbS$ and $\bbB = \bbE^{\Hr} \bbS+ \bbS \bbE$, to get
\begin{equation}
    \bbH(\hbS) - \bbH(\bbS) 
        = \sum_{k=0}^{\infty} h_{k} 
            \sum_{r=0}^{k-1} \left( 
                \bbS^{r} \bbE^{\Hr} \bbS^{k-r} + \bbS^{r+1} \bbE \bbS^{k-r-1} 
            \right)
        + \bbD
\end{equation}
with $\bbD$ such that $\|\bbD\| = \ccalO(\|\bbE\|_{2}^{2})$.

Next, we proceed to compute the output to an graph signal $\bbx$ with finite energy $\|\bbx\|<\infty$ which has a GFT given by $\tbx = [\tdx_{1},\ldots,\tdx_{N}]^{\Tr}$ so that
\begin{equation}
\bbx = \sum_{i=1}^{N} \tdx_{i} \bbv_{i}
\end{equation}
for $\{\bbv_{i}\}_{i=1}^{N}$ the eigenvector basis of the GSO $\bbS$. Then, we can compute
\begin{equation} \label{eqn:filterDifferenceWithX}
    \left[ \bbH(\hbS) - \bbH(\bbS) \right] \bbx 
        = \sum_{i=1}^{N} \tdx_{i} 
            \sum_{k=0}^{\infty} h_{k} 
                \sum_{r=0}^{k-1} \left( 
                    \bbS^{r} \bbE^{\Hr} \bbS^{k-r} 
                    + \bbS^{r+1} \bbE \bbS^{k-r-1} 
                \right) \bbv_{i}
            +\sum_{i=1}^{N} \tdx_{i} \bbD \bbv_{i}
\end{equation}
Let us consider first the product $\bbS^{r+1} \bbE \bbS^{k-r-1} \bbv_{i}$. It is immediate that $\bbS^{k-r-1}\bbv_{i} = \lambda_{i}^{k-r-1}\bbv_{i}$, so we focus on the product
\begin{equation}
    \bbE \bbv_{i} 
        = \sum_{n=1}^{N} m_{n} \bbu_{n} \bbu_{n}^{\Hr} \bbv_{i} 
        = m_{N} \sum_{n=1}^{N} 
            \frac{m_{n}}{m_{N}} \bbu_{n} \bbu_{n}^{\Hr} \bbv_{i}.
\end{equation}
The hypothesis that $\| \bbE/m_{N} - \bbI \| \leq \varepsilon$ is equivalent to $1 - \varepsilon \leq m_{n}/m_{N} \leq 1+\varepsilon$ for all $n = 1,\ldots,N$. Then, we can write $m_{n}/m_{N} = 1+ \delta_{n}$ with $|\delta_{n}| \leq \varepsilon$, which yields
\begin{equation} \label{eqn:Evi}
    \bbE \bbv_{i} 
        = m_{N} \bbv_{i} + m_{N} \bbw_{i} 
    \quad , \quad 
    \bbw_{i} 
        = \sum_{n=1}^{N} \delta_{n} \bbu_{n} \bbu_{n}^{\Hr} \bbv_{i}.
\end{equation}
Note that
\begin{equation} \label{eqn:wiBound}
    \| \bbw_{i} \|
        \leq \left\| \sum_{n=1}^{N} \delta_{n} \bbu_{n} \bbu_{n}^{\Hr} \right\| 
        \| \bbv_{i} \|
    = \max_{n=1,\ldots,N} |\delta_{n}| 
    \leq \varepsilon.
\end{equation}

Using \eqref{eqn:Evi} we get that
\begin{equation}
    \bbS^{r+1} \bbE \bbS^{k-r-1} \bbv_{i} 
        = m_{N} \lambda_{i}^{k} \bbv_{i} 
            + m_{N} 
                \bbV \lambda_{i}^{k-r-1} \bbLambda^{r+1} \bbV^{\Hr} 
                \bbw_{i}.
\end{equation}
And this can be used to compute
\begin{equation} \label{eqn:termE}
\begin{aligned}
    \sum_{k=0}^{\infty} h_{k} 
        \sum_{r=0}^{k-1} \bbS^{r+1} \bbE \bbS^{k-r-1} \bbv_{i} 
    & = m_{N} \sum_{k=0}^{\infty} h_{k} (k \lambda_{i}^{k} \bbv_{i})
        + m_{N} \bbV 
            \left( \sum_{k=0}^{\infty} h_{k} 
                \sum_{r=0}^{k-1} \lambda_{i}^{k-r-1} \bbLambda^{r+1} 
            \right) 
            \bbV^{\Hr} \bbw_{i} 
\\
    & = m_{N} \lambda_{i} h'(\lambda_{i}) \bbv_{i} 
        + m_{N} \bbV \diag(\hbg_{i}) \bbV^{\Hr} \bbw_{i}
\end{aligned}
\end{equation}
where vector $\hbg_{i} \in \reals^{N}$ is such that
\begin{equation}
[\hbg_{i}]_{j} = 
\sum_{k=0}^{\infty} h_{k} 
\sum_{r=0}^{k-1} \lambda_{i}^{k-r-1} \lambda_{j}^{r+1}.
\end{equation}
We note that if $j=i$ then $\lambda_{i}^{k-r-1} \lambda_{j}^{r+1} = \lambda_{i}^{k}$ and thus $[\hbg_{i}]_{i} = \lambda_{i} h'(\lambda_{i})$. For $j \neq i$, on the other hand, noting that $\sum_{r=0}^{k-1} \lambda_{i}^{k-r-1} \lambda_{j}^{r+1} = \lambda_{j} (\lambda_{i}^{k} - \lambda_{j}^{k})/(\lambda_{i}-\lambda_{j})$ we have
\begin{equation}
[\hbg_{i}]_{j} 
= \sum_{k=0}^{\infty} h_{k} 
\lambda_{j} 
\frac{\lambda_{i}^{k} - \lambda_{j}^{k}}{\lambda_{i} - \lambda_{j}}
= \frac{\lambda_{j}}{\lambda_{i}-\lambda_{j}} 
\sum_{k=0}^{\infty} h_{k} 
\left(\lambda_{i}^{k} - \lambda_{j}^{k} \right).
\end{equation}
Therefore,
\begin{equation} \label{eqn:hatg}
[\hbg_{i}]_{j} = 
\begin{cases}
\lambda_{i} h'(\lambda_{i}) 
& \text{if } j=i \\
\lambda_{j} 
\frac{h(\lambda_{i}) - h(\lambda_{j})}{\lambda_{i}-\lambda_{j}}
& \text{if } j \neq i
\end{cases}
\end{equation}
We also observe that $|[\hbg_{i}]_{j}| \leq G \lesssim \max\{C, 2B\}$ for all $j=1,\ldots,N$ due to the fact that $|h(\lambda)| \leq B$ and $|\lambda h'(\lambda)| \leq C$ due to the integral Lipschitz constraint.

We can get an expression analogous to \eqref{eqn:termE} for the term
\begin{equation} \label{eqn:termEH}
    \sum_{k=0}^{\infty} h_{k} 
        \sum_{r=0}^{k-1} \bbS^{r} \bbE^{\Hr} \bbS^{k-r} \bbv_{i}  
    = m_{N} \lambda_{i} h'(\lambda_{i})\bbv_{i} 
        + m_{N} \bbV \diag(\check{\bbg}_{i}) \bbV^{\Hr} \bbw_{i}
\end{equation}
where now
\begin{equation} \label{eqn:checkg}
[\check{\bbg}_{i}]_{j} = 
\begin{cases}
\lambda_{i} h'(\lambda_{i}) 
& \text{if } j=i \\
\lambda_{i} 
\frac{h(\lambda_{i}) - h(\lambda_{j})}{\lambda_{i}-\lambda_{j}}
& \text{if } j \neq i
\end{cases}
\end{equation}
where it also holds that $|[\check{\bbg}_{i}]_{j}| \leq G$.

Finally, using \eqref{eqn:termE} and \eqref{eqn:termEH} back in \eqref{eqn:filterDifferenceWithX}, and applying the norm, we get
\begin{align}
\left\| \left[ \bbH(\hbS) - \bbH(\bbS) \right] \bbx \right\|  
& \leq \left\| 2 m_{N} 
\sum_{i=1}^{N} \lambda_{i}h'(\lambda_{i}) \tdx_{i} \bbv_{i}  
\right\|
\label{eqn:firstOrderTerm} \\
& \quad + \left\| m_{N}  
\sum_{i=1}^{N} 
\bbV \diag(\hbg_{i}+ \check{\bbg}_{i}) \bbV^{\Hr} 
(\tdx_{i} \bbw_{i}) 
\right\|
\label{eqn:secondOrderTermEvi} \\
& \quad + \left\| \bbD \tbx \right\| 
\label{eqn:secondOrderTermTaylor}.
\end{align}
For the first order term \eqref{eqn:firstOrderTerm} we have
\begin{equation}
\left\| 2 m_{N} 
\sum_{i=1}^{N} 
\lambda_{i}h'(\lambda_{i}) 
\tdx_{i} \bbv_{i} 
\right\|^{2} 
= 4 |m_{N}|^{2} 
\sum_{i=1}^{N} 
|\lambda_{i} h'(\lambda_{i})|^{2}
|\tdx_{i}|^{2}
\end{equation}
since $\{\bbv_{i}\}_{i=1}^{N}$ form an orthonormal basis. Then, bounding $|m_{N}| \leq \varepsilon/2$ in virtue of $d(\bbS, \hbs) \leq \varepsilon/2$ and $|\lambda h'(\lambda)| \leq C$ for all $\lambda$, we get
\begin{equation} \label{eqn:boundFirstOrderTerm}
4 |m_{N}|^{2} 
\sum_{i=1}^{N} |\lambda_{i} h'(\lambda_{i})|^{2} |\tdx_{i}|^{2} 
\leq \varepsilon^{2} C^{2} 
\sum_{i=1}^{N} |\tdx_{i}|^{2} 
= \varepsilon^{2} C^{2} \| \bbx \|^{2}.
\end{equation}
For the second order term \eqref{eqn:secondOrderTermEvi} coming from $\bbE \bbv_{i}$, we have
\begin{equation}
 \left\| m_{N}  
\sum_{i=1}^{N}
\bbV \diag(\hbg_{i}+ \check{\bbg}_{i}) \bbV^{\Hr} 
(\tdx_{i} \bbw_{i}) \right\|_{2} 
 \leq |m_{N}|
\sum_{i=1}^{N} 
\| \bbV \diag(\hbg_{i}+ \check{\bbg}_{i}) \bbV^{\Hr}\|_{2}
|\tdx_{i} | \|\bbw_{i}\|_{2}
\end{equation}
where, by bounding $|m_{N}| \leq \varepsilon/2$, $\| \bbV \diag(\hbg_{i}+ \check{\bbg}_{i}) \bbV^{\Hr} \|_{2} \leq 2G$ in virtue of \eqref{eqn:hatg} and \eqref{eqn:checkg}, $\sum_{i=1}^{N} |\tdx_{i}| = \| \tbx \|_{1} \leq \sqrt{N} \|\tbx\|_{2} = \sqrt{N} \|\bbx\|_{2}$ and $\|\bbw_{i}\| \leq \varepsilon$ because of \eqref{eqn:wiBound}, we get
\begin{equation} \label{eqn:boundSecondOrderTermEvi}
\left\| m_{N}  
\sum_{i=1}^{N}
\bbV \diag(\hbg_{i}+ \check{\bbg}_{i}) \bbV^{\Hr}
(\tdx_{i} \bbw_{i}) 
\right\|
\leq \ccalO(\varepsilon^{2}) \| \bbx \|.
\end{equation}
Finally, for the second order term \eqref{eqn:secondOrderTermTaylor} stemming from the expansion of $\hbS^{k}$, we obtain 
\begin{equation} \label{eqn:boundSecondOrderTermTaylor}
\| \bbD \tbx \|_{2} \leq \ccalO(\| \bbE \|_{2}^{2}) \| \bbx \|_{2} 
\leq \ccalO(\varepsilon^{2}) \| \bbx \|_{2}.
\end{equation}
Using bounds \eqref{eqn:boundFirstOrderTerm}, \eqref{eqn:boundSecondOrderTermEvi} and \eqref{eqn:boundSecondOrderTermTaylor} back in \eqref{eqn:firstOrderTerm}, \eqref{eqn:secondOrderTermEvi} and \eqref{eqn:secondOrderTermTaylor}, respectively, we complete the proof.


\section{Proof of proposition \ref{prop:GSTcoeffStability}: GST coefficient stability}

We prove a more general case in which the low-pass average operator $U$ depends on the GSO $\bbS$ and is such that $\|U\| \leq B_{U}$ and $\|U(\bbS) - U(\hbS)\| \leq \varepsilon_{U}$. Prop.~\ref{prop:GSTcoeffStability} can be readily obtained from Prop.~\ref{prop:GSTcoeffStabilityWithU} below by setting $B_{U} = 1$ and $\varepsilon_{U}=0$ which is the case for the selected low-pass average operator $U=N^{-1} \bbone^{\Tr}$, that does not depend on $\bbS$.

%
%
\begin{proposition}[GST coefficient stability] \label{prop:GSTcoeffStabilityWithU}
    Let $\ccalG$ be a graph with GSO $\bbS$ and $\widehat{\ccalG}$ be the \emph{perturbed} graph with GSO $\hbS$, such that $d(\bbS, \hbS) \leq \varepsilon/2$. Let $\bbE \in \ccalE(\bbS, \hbS)$, consider its eigendecomposition $\bbE = \bbU \bbM \bbU^{\Hr}$ where the eigenvalues in $\bbM = \diag(m_{1},\ldots,m_{N})$ are ordered such that $|m_{1}| \leq \cdots \leq |m_{N}|$, and assume that the structural constraint $\| \bbE / m_{N} - \bbI \| \leq \varepsilon$ holds. Consider a GST with $L$ layers and $J$ wavelet scales $h_{j}(\lambda)$, each of which satisfies the integral Lipschitz constraint $|\lambda h_{j}'(\lambda)| \leq C$ and conform a frame with bounds $0 < A \leq B$ [cf. \eqref{eqn:waveletFrame}]. Then, for the coefficient $\phi_{p_{j}(\ell)}$ associated to path $p_{j}(\ell) = (j_{1},\ldots,j_{\ell})$ it holds that
    \begin{equation}
        | \phi_{p_{j}(\ell)} (\bbS, \bbx) - \phi_{p_{j}(\ell)} (\hbS, \bbx) | \leq \left( \varepsilon_{U} B^{\ell} + B_{U} \varepsilon C \ell B^{\ell-1} \right) \| \bbx \|
    \end{equation}
\end{proposition}
%

Starting with \eqref{eqn:scatteringCoefficients}, using graph convolutions \eqref{eqn:graphConvolution} and recalling that we can write $\bbh_{j} \ast_{\bbS} \bbx = \bbH_{j}(\bbS) \bbx$, we get
\begin{equation}
\begin{aligned}
\left| \phi_{p_{j}(\ell)}(\bbS, \bbx) - \phi_{p_{j}(\ell)}(\hbS, \bbx)\right| 
 & = \left| U(\bbS) (\rho \bbH_{j}(\bbS))_{p_{j}(\ell)} \bbx
- U(\hbS) (\rho \bbH_{j}(\hbS))_{p_{j}(\ell)} \bbx \right| \\
&  \leq \left| U(\bbS) (\rho \bbH_{j}(\bbS))_{p_{j}(\ell)} \bbx 
- U(\hbS) (\rho \bbH_{j}(\bbS))_{p_{j}(\ell)} \bbx \right| \\
&   \qquad  + \left| U (\hbS) (\rho \bbH_{j}(\bbS))_{p_{j}(\ell)} \bbx 
- U (\hbS) (\rho \bbH_{j}(\hbS))_{p_{j}(\ell)} \bbx \right|
\end{aligned}
\end{equation}
where we have added and subtracted $U(\hbS) (\rho h_{j}(\bbS))_{p_{j}(\ell)}\bbx$, and then applied the triangle inequality. Applying Cauchy-Schwarz inequality to each term, we get
\begin{equation} \label{eqn:targetBoundCoeff}
\begin{aligned}
 \left| \phi_{p_{j}(\ell)}(\bbS, \bbx) - \phi_{p_{j}(\ell)}(\hbS, \bbx)\right|
&   \leq \| U(\bbS) - U(\hbS) \| 
\| (\rho \bbH_{j}(\bbS))_{p_{j}(\ell)}\bbx \| \\
& \quad + \| U(\hbS) \| 
\| (\rho \bbH_{j}(\bbS))_{p_{j}(\ell)}\bbx 
- (\rho \bbH_{j}(\hbS))_{p_{j}(\ell)}\bbx \|.
\end{aligned}
\end{equation}
We proceed by bounding, one by one, these four terms. The first one, is bounded by hypothesis
\begin{equation} \label{eqn:firstTerm}
\| U(\bbS) - U(\hbS) \| \leq \varepsilon_{U}. 
\end{equation}
For the second term, we recall that the nonlinearity is nonexpansive, i.e. $\|\rho\| \leq 1$, and use the definition of operator norm together with the property of submultiplicativity
\begin{equation}
\| (\rho \bbH_{j}(\bbS))_{p_{j}(\ell)}\bbx \| 
\leq \| \rho \bbH_{j_{\ell}}(\bbS) \| 
\cdots 
\| \rho \bbH_{j_{1}}(\bbS) \| 
\| \bbx \|
\end{equation}
and thus, together with the frame condition \eqref{eqn:waveletFrame}, we obtain
\begin{equation} \label{eqn:secondTerm}
\| (\rho \bbH_{j}(\bbS))_{p_{j}(\ell)}\bbx \| 
\leq B^{\ell} \| \bbx \|.
\end{equation}
The third term is bounded by the hypothesis that the summarizing linear operator is bounded
\begin{equation} \label{eqn:thirdTerm}
\| U(\hbS) \| \leq B_{U}.
\end{equation}
The fourth and last term is slightly more involved. We can bound it in a recursive fashion as follows. First, add and subtract $\rho \bbH_{j_{\ell}}(\bbS) \rho \bbH_{j_{\ell-1}}(\hbS) \cdots \rho \bbH_{j_{1}}(\hbS)$ and use the triangle inequality to obtain
\begin{align} \label{eqn:fourthTerm}
& \| (\rho \bbH_{j}(\bbS))_{p_{j}(\ell)}\bbx 
- (\rho \bbH_{j}(\hbS))_{p_{j}(\ell)}\bbx \| \nonumber \\
& \qquad \leq \| \rho \bbH_{j_{\ell}}(\bbS) 
\left( \rho \bbH_{j_{\ell-1}}(\bbS) \cdots \rho \bbH_{j_{1}}(\bbS)
- \rho \bbH_{j_{\ell-1}}(\hbS) \cdots \rho \bbH_{j_{1}}(\hbS) \right)
\| \nonumber \\
& \qquad \qquad + \| 
\left( \rho \bbH_{j_{\ell}}(\bbS) - \rho \bbH_{j_{\ell}}(\hbS)\right)
\rho \bbH_{j_{\ell-1}}(\hbS) \cdots \rho \bbH_{j_{1}}(\hbS) 
\|.
\end{align}
Now, using submultiplicativity and defining
\begin{equation}
\mathrm{bound}(\ell) = \| \rho \bbH_{j_{\ell}}(\bbS) \cdots \rho \bbH_{j_{1}}(\bbS)
- \rho \bbH_{j_{\ell}}(\hbS) \cdots \rho \bbH_{j_{1}}(\hbS)\|
\end{equation}
we observe that \eqref{eqn:fourthTerm} becomes the recursive inequality
\begin{equation} \label{eqn:recursiveInequality}
\mathrm{bound}(\ell) 
\leq B \ \mathrm{bound}(\ell-1) 
+ \varepsilon C B^{\ell-1}
\end{equation}
where we have used that $\|\rho \bbH_{j_{\ell}}(\bbS)\| \leq B$ by the frame condition, that $\|\bbH_{j}(\bbS) - \bbH_{j}(\hbS)\| \leq \varepsilon C$ due to Prop.~\ref{prop:waveletStability}, and that $\| \rho \bbH_{j_{\ell-1}}(\hbS) \cdots \rho \bbH_{j_{1}}(\hbS)  \| \leq B^{\ell-1}$ by the same submultiplicativity and frame argument of \eqref{eqn:secondTerm}.

Solving the recursive inequality in \eqref{eqn:recursiveInequality} we reach
\begin{equation}
\mathrm{bound}(\ell) 
\leq B^{\ell-1} \ \mathrm{bound}(1) 
+ (\ell-1) \varepsilon C B^{\ell-1}
\end{equation}
and noting that
\begin{equation}
\mathrm{bound}(1) 
= \| \rho \bbH_{j_{1}}(\bbS) - \rho \bbH_{j_{1}}(\hbS) \| 
\leq \varepsilon C
\end{equation}
by Prop.~\ref{prop:waveletStability} we finally bound the fourth term \eqref{eqn:fourthTerm} by
\begin{equation} \label{eqn:fourthTermBound}
\| (\rho \bbH_{j}(\bbS))_{p_{j}(\ell)}\bbx 
- (\rho \bbH_{j}(\hbS))_{p_{j}(\ell)} \bbx \| 
\leq \ell \varepsilon C B^{\ell-1}
\end{equation}
Finally, substituting \eqref{eqn:firstTerm}, \eqref{eqn:secondTerm}, \eqref{eqn:thirdTerm} and \eqref{eqn:fourthTermBound} back in \eqref{eqn:targetBoundCoeff}, we complete the proof.


\section{Proof of theorem \ref{thm:GSTstability}: GST stability}

In this case, we also prove a more general case for a low-pass average operator $U$ that depends on the GSO $\bbS$ and is such that $\|U\| \leq B_{U}$ and $\|U(\bbS) - U(\hbS)\| \leq \varepsilon_{U}$. Theorem~\ref{thm:GSTstability} can be readily obtained from Theorem~\ref{thm:GSTstability} below by setting $B_{U}=1$ and $\varepsilon_{U}=0$ which is the case for the selected low-pass average operator $U=N^{-1} \bbone^{\Tr}$, that does not depend on $\bbS$.

%
%
\begin{theorem}[GST stability] \label{thm:GSTstabilityWithU}
    Let $\ccalG$ be a graph with GSO $\bbS$ and $\widehat{\ccalG}$ be the \emph{perturbed} graph with GSO $\hbS$, such that $d(\bbS, \hbS) \leq \varepsilon/2$. Let $\bbE \in \ccalE(\bbS, \hbS)$, consider its eigendecomposition $\bbE = \bbU \bbM \bbU^{\Hr}$ where the eigenvalues in $\bbM = \diag(m_{1},\ldots,m_{N})$ are ordered such that $|m_{1}| \leq \cdots \leq |m_{N}|$, and assume that the structural constraint $\| \bbE / m_{N} - \bbI \| \leq \varepsilon$ holds. Consider a GST with $L$ layers and $J$ wavelet scales $h_{j}(\lambda)$, each of which satisfies the integral Lipschitz constraint $|\lambda h_{j}'(\lambda)| \leq C$ and conform a frame with bounds $0 < A \leq B$ [cf. \eqref{eqn:waveletFrame}]. Then, it holds that
    \begin{equation} \label{eqn:GSTcoeffStabilityWithU}
    \left\| \bbPhi(\bbS, \bbx) - \bbPhi (\hbS, \bbx) \right\| \leq 
        \left[
        \varepsilon_{U}^{2} \xi^{(0)}_{BJL}
        + 2 \varepsilon_{U} B_{U} \frac{\varepsilon C}{B} \xi^{(1)}_{BJL}
        + B_{U}^{2} \left(\frac{\varepsilon C}{B}\right)^{2} \xi^{(2)}_{BJL}
        \right]^{1/2} \| \bbx\|
    \end{equation}
    with $\xi^{(r)}_{BJL} = \sum_{\ell=0}^{L-1} \ell^{r} (B^{2}J)^{\ell}$.
\end{theorem}
%

From \eqref{eqn:energyScattering}, we get
\begin{equation}
    \| \bbPhi(\bbS, \bbx) - \bbPhi(\hbS, \bbx) \|^{2} 
        = \sum_{\ell=0}^{L-1} 
            \sum_{j=1}^{J^{\ell}} 
                |\phi_{p_{j}(\ell)}(\bbS, \bbx) 
                    - \phi_{p_{j}(\ell)}(\hbS, \bbx)|^{2}.
\end{equation}
Now, each term in the sum, can be bounded by means of Prop.~\ref{prop:GSTcoeffStability}, so that
\begin{equation}
    \| \bbPhi(\bbS, \bbx) - \bbPhi(\hbS, \bbx) \|^{2} 
        \leq \sum_{\ell=0}^{L-1} 
            \sum_{j=1}^{J^{\ell}} 
                \left(
                    \varepsilon_{U} B^{\ell} \| \bbx \|
                    + B_{U} \varepsilon C \ell B^{\ell-1} \| \bbx\|
                \right)^{2}.
\end{equation}
Expanding the square, and taking $\|\bbx\|^{2}$ out of the sum, yields
\begin{equation}
    \| \bbPhi(\bbS, \bbx) - \bbPhi(\hbS, \bbx) \|^{2} 
        \leq \| \bbx \|^{2} \sum_{\ell=0}^{L-1} 
            \sum_{j=1}^{J^{\ell}}
                \left(
                    \varepsilon_{U}^{2} B^{2\ell}
                    + 2\varepsilon_{U} B_{U}\varepsilon C \ell B^{2\ell-1}
                    + B_{U}^{2} \varepsilon^{2} C^{2} \ell^{2} B^{2(\ell-1)}
                \right).
\end{equation}
We note that no term in the inner sum depends on $j$, so we obtain
\begin{align}
    \| \bbPhi(\bbS, \bbx) &- \bbPhi(\hbS, \bbx) \|^{2} 
        \leq \| \bbx \|^{2} \sum_{\ell=0}^{L-1} 
            J^{\ell}
            \left(
                \varepsilon_{U}^{2} B^{2\ell}
                + 2\varepsilon_{U} B_{U}\varepsilon C \ell B^{2\ell-1}
                + B_{U}^{2} \varepsilon^{2} C^{2} \ell^{2} B^{2(\ell-1)}
            \right) \nonumber \\
        & \leq \| \bbx \|^{2} \sum_{\ell=0}^{L-1} 
            \left(
                \varepsilon_{U}^{2} (JB^{2})^{\ell} 
                + 2\varepsilon_{U} B_{U}(\varepsilon C/B) \ell (JB^{2})^{\ell}
                + B_{U}^{2} (\varepsilon C/B)^{2} \ell^{2} (JB^{2})^{\ell}
        \right).
\end{align}
Assuming $JB^{2} \neq 1$, we can use the geometric sum to get
\begin{equation}
    \| \bbPhi(\bbS, \bbx) - \bbPhi(\hbS, \bbx) \|^{2}
        \leq \| \bbx\|^{2} \left[
            \varepsilon_{U}^{2} \xi^{(0)}_{BJL}
            + 2 \varepsilon_{U} B_{U} \frac{\varepsilon C}{B} \xi^{(1)}_{BJL}
            + B_{U}^{2} \left(\frac{\varepsilon C}{B}\right)^{2} \xi^{(2)}_{BJL}
        \right]
\end{equation}
with
\begin{align}
    \xi^{(0)}_{BJL} & = \sum_{\ell=0}^{L-1} (B^{2}J)^{\ell}
        = \frac{(B^{2}J)^{L}-1}{B^{2}J - 1} \\
    \xi^{(1)}_{BJ L} & = \sum_{\ell=0}^{L-1} \ell(B^{2}J)^{\ell}
        = \frac{ B^{2}J + (L-1) (B^{2}J)^{L-1} - L (B^{2}J)^{L} }{ (B^{2}j-1)^{2} } \\
    \xi^{(2)}_{BJ L} & =  \sum_{\ell=0}^{L-1} \ell^{2} (B^{2}J)^{\ell}  \\
        & =
          \frac{ (1+2L - 2L^{2}) (B^{2}J)^{L+1} + L^{2} (B^{2}J)^{L} + (L-1)^{2} (B^{2}J)^{L+2} - (B^{2}J)^{2} - (BJ)^{2} }{ (B^{2}J -1)^{3} }  \nonumber
\end{align}
Finally, we apply the square root to complete the proof.


\section{Details on numerical experiments}

Experiment \ref{subsec:smallWorld} is a synthetic experiment where we can exercise full control on the perturbation size $\varepsilon$ [cf. \eqref{eqn:perturbationSize}]. The objective is to show how stable is the GST when compared to the GFT, and also to show how tight the bound is. Experiments \ref{subsec:author} and \ref{subsec:facebook} are based on real-world data, in problem formulations analogous to \cite{Gama19-Scattering}. The objective is to show that the GST is a useful representation, yielding similar performance than the GFT (i.e. that they capture, at least, as rich information as the GFT). Additionally, we show how stable the GST is to real-world perturbations (i.e. perturbations that are not synthetically controlled by fixing $\varepsilon$).

We consider three different GSTs. In all cases, we consider $J=6$ scales and $L=3$ layers yielding $43$ coefficients. First, we consider the use of a monic cubic polynomial as the generating kernel, see \cite[eq. (65)]{Hammond11-Wavelets} and ensuing discussion for details. We set $x_{1}$ to be $\lambda_{\lfloor N/4 \rfloor}$ and $x_{2} = \lambda_{\lceil 3N/4 \rceil}$ (i.e. for the eigenvalues in increasing order, the first fourth of the eigenvalues are affected by the monic polynomial $x_{1}^{-\alpha}x^{\alpha}$, and for the last fourth, by $x_{2}^{\beta} x^{-\beta}$). The values of $\alpha=\beta=2$ and $K=20$ are the same as in \cite{Hammond11-Wavelets}. The cubic polynomial for the eigenvalues located between $x_{1}$ and $x_{2}$  is designed so that the Kernel has continuous first derivatives. The adopted GSO for the monic cubic polynomial GST is the normalized Laplacian $\bbS = \bbD^{-1/2}(\bbD - \bbW) \bbD^{-1/2}$ with $\bbD = \diag(\bbW \bbone)$ the degree matrix and $\bbW$ the adjacency matrix, as suggested in \cite{Hammond11-Wavelets}. We denote this GST as $\bbPhi_{\text{MC}}(\bbS, \bbx)$ and refer to it as ``Monic Cubic''.

Second, we employ a tight Hann wavelet kernel, see \cite[Example 1]{Shuman15-Wavelets} for details. More specifically, we implement a generating kernel in \cite[eq. (9)]{Shuman15-Wavelets} with $K=1$, $a_{0}=a_{1}=1/2$, and $R=3$. We do warping as suggested in \cite[Sec.~IV]{Shuman15-Wavelets} with a warping function $\omega(\lambda) = \log(\lambda)$. We then construct the remaining wavelets from the generating kernel as in \cite[eq. (12)]{Shuman15-Wavelets} with a scaling function given by \cite[eq. (13)]{Shuman15-Wavelets}. The adopted GSO is also the normalized Laplacian. We denote this GST as $\bbPhi_{\text{TH}}(\bbS, \bbx)$ and refer to it as ``Tight Hann''.

Third, we compare with the graph diffusion scattering of \cite{Gama19-Scattering}. In this case, each wavelet is obtained as $\bbH_{j} = \bbT^{2^{j-1}} (\bbI - \bbT^{2^{j-1}})$, $j=1,\ldots,J$ for $\bbT = 1/2 (\bbI + \bbD^{-1/2} \bbW \bbD^{-1/2})$ the lazy diffusion operator, which we adopt as the GSO. The low-pass average operator $U$ in this case is not the constant $N^{-1} \bbone^{\Tr}$ but the suggested $U = \bbd / \| \bbd\|_{1}$ (i.e. an average by the degree of each node). We denote this GST as $\bbPhi_{\text{D}}(\bbS, \bbx)$ and refer to it as ``Diffusion''.

To compare the stability of the graph-based representation given by the GST, we construct another graph-based representation, namely, the graph Fourier transform (GFT). Given a GSO $\bbS = \bbV \bbLambda \bbV^{\Hr}$ and a graph signal $\bbx$, the GFT is computed as $\tbx(\bbS,\bbx) = \bbV^{\Hr} \bbx$, where the dependence on $\bbS$ comes through $\bbV$. We choose the normalized Laplacian as the GSO for computing the eigenbasis $\bbV$. We note that, unlike the GST, the number of coefficients in the GFT representation is $N$. Therefore, for fair comparison, in experiments \ref{subsec:author} and \ref{subsec:facebook} we select a number of GFT coefficients equal to the number of GST coefficients. We denote the GFT as $\tbx(\bbS, \bbx)$ and refer to it as ``GFT''.


\subsection{Relative representation error: Small world graphs} \label{subsec:smallWorld}

In this first experiment, we consider a small world graph of $N=100$ nodes, generated randomly by using an edge probability $p_{\text{SW}} = 0.5$ and a rewiring probability $q_{\text{SW}} = 0.1$. We then consider a white noise signal $\bbx$ with power $\sigma_{x}^{2}$ defined on top of this graph, and compute the corresponding representations $\bbPhi(\bbS, \bbx)$ for all three GSTs and $\tbx(\bbS, \bbx)$ for the GFT. We consider perturbations of the adjacency matrix $\bbW$ given by $\hbW = \bbW + \bbE^{\Hr} \bbW + \bbW \bbE$ [cf. \eqref{eqn:errorMatrices}], where error matrix $\bbE = \diag(\bbe)$ is a diagonal matrix with $\bbe$ being uniformly random, chosen such that $\|\bbE\| \leq \varepsilon/2$ and $\|\bbE/e_{\max} - \bbI \| \leq \varepsilon$ for $e_{\max} = \sign\{\argmax |[\bbe]_{n}|\} \max |[\bbe]_{n}|$. Such a deformation amounts for a local dilation of the edge weights (i.e. the edge weights of the neighborhood of each node are dilated by different values). We control the value of $\varepsilon$ as a parametric sweep from $0.1 \cdot \sigma_{x}^{2}$ to $1 \cdot \sigma_{x}^{2}$.

To account for the different sources of randomness, we generate $10$ random connected graph realizations of the small world model, and for each of these $10$ graphs we sweep for $10$ different values of $\varepsilon$ linearly spaced. For each value of $\varepsilon$ we consider $10$ random realizations of the error matrix $\bbE$, and for each of these perturbations, we simulate $1000$ test signals $\bbx$ assumed to be white gaussian with zero-mean and power $\sigma_{x}^{2} = 1$. We compute $\bbPhi(\bbS,\bbx)$ and $\bbPhi(\hbS, \bbx)$ for each of the three different GSTs and also $\tbx(\bbS,\bbx)$ and $\tbx(\hbS, \bbx)$ for the GFT. We calculate $\| \bbPhi(\bbS, \bbx) - \bbPhi(\hbS, \bbx)\|/\|\bbPhi(\bbS, \bbx)\|$ for each of the signal and average across all $1000$ test signals, and then average these means across all $10$ random realizations of the error matrix to obtain an estimate of the relative representation error for each graph, for each value of $\varepsilon$. We proceed analogously for the GFT. For each value of $\varepsilon$ we get $10$ estimates fo the relative representation error, one for each random graph realization. We average these across the $10$ graphs and plot them as solid lines of Fig.~\ref{sw}. We estimate the standard deviation across the $10$ graphs and plot them as the error bars. We also show, in dashed line, the value of the bound \eqref{eqn:GSTcoeffStability} for the GST using the monic cubic polynomial wavelets (we choose this one due to its simplicity in computing the frame bound $B$ and the integral Lipschitz constant $C$).


\subsection{Authorship attribution: Jane Austen} \label{subsec:author}

In this experiment, we consider the problem of authorship attribution. The objective is that, given a text excerpt, we can accurately attribute it to a given author. In particular, we consider works authored by Jane Austen, in the same setting as in \cite{Gama19-Scattering}. To cast this problem as a graph signal classification problem, we proceed as follows. Given a training set of text excerpts (i.e. text excerpts that we know have been authored by Jane Austen), we build a word adjacency network (WAN) using functional words (i.e. words without semantic meaning such as connectors) by determining their relative positioning in the text. It has been noted that the relative positioning of functional words offers a stylometric signature of the author, see \cite{Segarra15-WANs} for details. Once the graph is built with $N$ functional words, we ensure it is connected and make it undirected by recomputing the edge weights to be the average of the incoming and outgoing edge weights. Each of these functional words act as a node in the network. We can then associate a graph signal (on top of this WAN graph) to each text by counting the frequency of appearance of the functional words. It is then expected that if the frequency of functional words bears strong relation with the graph, then the given text was written by the author for which the WAN was built.

We consider $N=224$ function words, and a corpus of $771$ text fragments (of approximately $1,000$ words) authored by Jane Austen. We split at random this corpus in training, validation and test sets, and use the training set to build the WAN. It is important to note that the texts included in the training set are the only ones used to build the WAN graph, and therefore, the graph is different depending on what texts were selected for the training set. This is a realistic scenario that models the perturbation in the underlying support arising from an estimation of the graph topology (i.e. we do not know the specific graph topology, but estimate it from data, and therefore the true graph topology might be different from the one we are actually using). We consider $10$ different split ratios ranging linearly from $0.2$ to $0.9$ which implies that the number of texts used to build the WAN varies from $154$ to $694$.

Once we build the WAN graph, we use a linear SVM to classify the graph signals. To train this SVM we use the same texts included in the training set (labeled as $1$ since they were written by the author of interest), and we add to the training set an equivalent number of texts written by other contemporary authors (labeled as $0$ since they were written by other authors, such as Emily Bront\"{e}, Edgar Allan Poe, Charles Dickens, among many others). We compute the relevant representation $\bbPhi$ and $\tbx$ for each text in the training set and use this representation to train $4$ different SVMs (one for each of the three GSTs and one for the GFT). To use representations of same size, we consider the GFT to project the signal only on the first $43$ eigenvectors of the GSO (low-pass filter). We then build a test set with the remaining texts by the author (those not used to build the WAN nor to train the SVM) and add an equivalent number of texts by other contemporary authors. The classification accuracy at test time for different sizes of training set is shown in Fig.~\ref{wan}.

To account for randomness, for each split ratio (i.e. for each total number of training samples), we simulate $20$ different data splits. We compute the classification accuracy averaged over the test set for each of these $20$ different data splits. By averaging over the $20$ data splits we obtain the mean classification accuracy showed in solid lines in Fig.~\ref{wan}. We also include the standard deviation estimated from these $20$ data splits. While the richness of the GST representation can be observed by the fact that the classification accuracy is comparable to that achieved by using the GFT representation, we can also observe the stability of the GST when compared to the GFT. More specifically, we see (i) that the mean value of the classification accuracy of the GFT oscillates much more than the mean value of the GST, and (ii) that the error bars for the GFT are much larger than those for the GST. This shows that, depending on how we build the unerlying graph (i.e. which and how many texts we use to estimate the WAN), the classification accuracy by using the GFT representation can vary wildly.


\subsection{Source localization: Facebook subnetwork} \label{subsec:facebook}

For the final example, we consider a source localization problem (synthetic data) on a Facebook subnetwork of $N=234$ users (real-world data), the same as in \cite{Gama19-Scattering}. This Facebook graph, exhibiting a two-community topology, is a subnetwork of the larger $4,039$ user graph provided in \cite{McAuley12-EgoNets}. The problem of source localization consists in observing a diffusion signal and pinpointing to where it started. In the context of graph signals, we consider a signal $\bbdelta_{c}$ which is a signal with a $1$ at node $c$ and $0$ elsewhere. Then, we observed the diffused signal $\bbx = \bbW^{t} \bbdelta_{c}$ for some unknown time $t < t_{\max}$ and we want to estimate the community $c$ that originated the rumor. In a two-community graph, this is a binary classification problem. This problem is analogous to identifying the source of a rumor that spread through the social network.

In this case, we consider perturbations stemming from randomly dropping edges with probability $p$, ranging in $10$ logarithmically-spaced points from $0.01$ to $0.3$, as in \cite{Gama19-Scattering}. This models changing friendships in the network. We use the underlying given graph as $\bbS$ which we use to build the representations $\bbPhi(\bbS,\bbx)$ and $\tbx(\bbS,\bbx)$, but we use data $\bbx$ generated on graph $\hbS$ that corresponds to some random realization of the edge dropping. Again, the objective of this simulation is twofold: (i) to show that using the GST representation achieves as good classification accuracy as using the GFT, and (ii) that the GST is more stable than the GFT. We consider $43$ GFT coefficients belonging to middle frequencies (bandpass filter).

To perform the classification, we train a Linear SVM on the representations obtained for each of the three GSTs and the GFT, analogously to experiment \ref{subsec:author}. We train the SVMs by generating $1,000$ training samples $\bbx = \bbW^{t} \bbdelta_{c}$ for $c \in \{c_{0}, c_{1}\}$ and random $t < t_{\max} = 20$. The source nodes $c_{0}$ and $c_{1}$ are the nodes numbered $38$ and $224$ since each of them belongs to a different community, and half of the training samples were originated at $c_{0}$ and the other half at $c_{1}$. For testing, we generate $200$ new samples, half for each community, with random diffusion times $t < t_{\max}$. Results are shown in Fig.~\ref{fb}.

To account for randomness, we generate $20$ different random edge-failing graph realizations, for each value of $p$ simulated. We average across these $20$ realizations to obtain the solid lines in Fig.~\ref{fb}, and compute the standard deviation for the error bars. We observe that $\bbPhi_{\text{D}}$ and $\bbPhi_{\text{TH}}$ perform consistently better (on average) than $\bbPhi_{\text{MC}}$ and $\tbx$. In terms of stability, both $\bbPhi_{\text{D}}$ and $\bbPhi_{\text{TH}}$ are more stable than $\bbPhi_{\text{MC}}$ and $\tbx$, but it is $\tbx$ that exhibits the largest error bars. The fact that, in this experiment, the GST representation seems to be less stable than in experiment \ref{subsec:author}, can be explained from the fact that dropping an edge incurs in a huge cost in terms of the structural constraint $\| \bbE/m_{N} - \bbI \|$.